\documentclass[a4paper,fleqn]{cas-sc}

\ExplSyntaxOn
\cs_gset:Npn \__first_footerline:
  {
    \group_begin:
    \small
    \sffamily
    \__short_authors: 
    \group_end:
  }
\ExplSyntaxOff

\usepackage[numbers]{natbib}
\usepackage{mathrsfs}
\usepackage{graphicx}%
\usepackage{multirow}%
\usepackage{amsmath,amssymb,amsfonts}%
\usepackage{amsthm}%
\usepackage{mathrsfs}%
\usepackage[title]{appendix}%
\usepackage{xcolor}%
\usepackage{textcomp}%
\usepackage{manyfoot}%
\usepackage{booktabs}%
\usepackage{algorithm}%
\usepackage{algorithmicx}%
\usepackage{algpseudocode}%
\usepackage{listings}%
\usepackage{array}
\usepackage{tikz}
\usetikzlibrary{positioning}
\usepackage{tabularx}
\usepackage{pifont}

\usepackage{lscape}  % or \usepackage{pdflscape}
\usepackage{colortbl} % for \rowcolor and \cellcolor
\usepackage[table,xcdraw]{xcolor} % for colored cells

\usepackage{float}

\usepackage[english]{babel} % In preamble

\definecolor{lightgray}{gray}{0.94}
\definecolor{lightblue}{RGB}{224,243,255}
\definecolor{lightgreen}{RGB}{235,255,235}
\definecolor{lightpink}{RGB}{255,239,239}
\definecolor{lightyellow}{RGB}{255,253,235}

\definecolor{headerblue}{RGB}{208,225,242} % 表头蓝
\definecolor{rowblue}{RGB}{240,246,251}    % 斑马纹浅蓝

\usepackage{pdflscape} % 放在导言区

\DeclareUnicodeCharacter{202F}{\,}

\theoremstyle{thmstyleone}%
\theoremstyle{thmstyletwo}%

\theoremstyle{thmstylethree}%

\def\tsc#1{\csdef{#1}{\textsc{\lowercase{#1}}\xspace}}
\tsc{WGM}
\tsc{QE}
\tsc{EP}
\tsc{PMS}
\tsc{BEC}
\tsc{DE}
\begin{document}
\let\WriteBookmarks\relax
\def\floatpagepagefraction{1}
\def\textpagefraction{.001}

% \shorttitle{Reasoning in Computer Vision}
\shorttitle{Reasoning in Computer Vision: Taxonomy, Models, Tasks, and Methodologies}

\shortauthors{Sarkar et~al.}

% \title [mode = title]{This is a specimen $a_b$ title}                  
\title[mode = title]{Reasoning in Computer Vision: Taxonomy, Models, Tasks, and Methodologies}

% \tnotemark[1,2]

% \tnotetext[1]{This document is the results of the research
%    project funded by the National Science Foundation.}

% \tnotetext[2]{The second title footnote which is a longer text matter
%    to fill through the whole text width and overflow into
%    another line in the footnotes area of the first page.}


% \author[1]{\fnm{Ayushman} \sur{Sarkar}}\email{ayushmansarkar123@gmail.com}

% \author[2]{\fnm{Mohd Yamani Idna} \sur{Idris}}\email{yamani@um.edu.my}

% \author*[2]{\fnm{Zhenyu} \sur{Yu}}\email{yuzhenyuyxl@foxmail.com}

% \affil[1]{\orgdiv{Department of Computer Science and Engineering}, \orgname{Birbhum Institute of Engineering and Technology}, \orgaddress{\city{Suri}, \postcode{731101}, \state{West Bengal}, \country{India}}}

% \affil*[2]{\orgdiv{Faculty of Computer Science and Information Technology}, \orgname{Universiti Malaya}, \orgaddress{\city{Kuala Lumpur}, \postcode{50603}, \country{Malaysia}}}

\author[1]{Ayushman Sarkar}
\ead{ayushmansarkar123@gmail.com}
\affiliation[1]{organization={Department of Computer Science and Engineering, Birbhum Institute of Engineering and Technology},
    city={Suri},
    postcode={731101},
    state={West Bengal},
    country={India}}

% Third author (corresponding author)
\author[2]{Zhenyu Yu}
\cormark[1]
\ead{yuzhenyuyxl@foxmail.com}
\affiliation[2]{organization={College of Computer Science and Artificial Intelligence, Fudan University},
    city={Shanghai},
    postcode={200433},
    country={China}}
    
\author[3]{Mohd Yamani Idna Idris}
% \cormark[1]
\ead{yamani@um.edu.my}
\affiliation[3]{organization={Faculty of Computer Science and Information Technology, Universiti Malaya},
    city={Kuala Lumpur},
    postcode={50603},
    country={Malaysia}}

\cortext[cor1]{Corresponding author}

\begin{abstract}
Visual reasoning matters for many computer vision tasks that go beyond surface-level object detection and classification. Despite progress in relational, symbolic, temporal, causal, and commonsense reasoning, existing surveys typically cover only one part of the problem, such as visual question answering, scene-graph generation, neuro-symbolic AI, or multimodal chain-of-thought, and rarely analyze reasoning types, methodologies, and evaluation protocols together. This survey addresses that gap. Following a structured literature review, we group visual reasoning into five major types (relational, symbolic, temporal, causal, and commonsense) and examine how each is implemented across methods that range from graph-based models, memory networks, attention mechanisms, and neuro-symbolic systems to reasoning with vision–language models (VLMs) and multimodal large language models (MLLMs), including visual chain-of-thought, visual programming, and tool-augmented and test-time reasoning. We then review evaluation protocols for functional correctness, structural consistency, and causal validity, and we analyze their limits in generalizability, reproducibility, faithfulness, and explanatory power. We also identify open challenges: scaling to complex scenes, integrating symbolic and neural paradigms more deeply, the shortage of comprehensive benchmarks, language-prior shortcuts and hallucination in foundation models, and reasoning under weak supervision. Finally, we set out a research agenda for vision systems and argue that connecting perception and reasoning is necessary for transparent, trustworthy, and cross-domain models, especially in high-stakes settings such as autonomous driving and medical diagnostics.
\end{abstract}

% \includegraphics{figs/grabs.pdf}

% % Research highlights
% \begin{highlights}
% \item Unifies relational, symbolic, temporal, causal and commonsense visual reasoning.
% \item Spans classical graph and neuro-symbolic models to VLM/MLLM-based reasoning.
% \item Covers visual chain-of-thought, visual programming and test-time reasoning.
% \item Reviews functional, structural and causal evaluation of reasoning faithfulness.
% \item Maps benchmark gaps and outlines a research agenda for trustworthy systems.
% \end{highlights}

% Keywords

\maketitle

\section{Introduction}

\label{sec:introduction}

Computer vision has advanced rapidly over the past decade, moving from perception-driven applications such as image classification and object detection to harder problems such as scene understanding, relational modeling, and VQA \cite{zellers2019recognition,cao2021attention,khan2022transformers,shen2025gfsnet,rossi2025advances}. Yet how to give vision systems \textbf{genuine reasoning capability} remains unsolved. Visual reasoning extends well beyond object recognition: it requires inference over object attributes, relations, and causal dependencies in space and time \cite{yang2022causalvqa,li2023weakly,shen2025gfsnet,vosoughi2024cross,chen2025role}. It links low-level perception with high-level cognition. Progress has been substantial, but clear gaps remain in explainability, cross-domain generalization, and the handling of complex tasks.

These limitations are more severe in safety-critical applications such as autonomous vehicles, clinical diagnosis, and satellite-based environmental monitoring, where transparency and credibility directly affect deployment \cite{palikhe2025towards, park2020visualcomet,sado2023explainable,vouros2022explainable,yang2021unbox,yeo2025comprehensive}. Although deep networks have strong predictive power, their internal reasoning is opaque \cite{wu2022multi, zhang2021vinvl,lee2024unlocking}. Existing explainable AI (XAI) methods such as Grad-CAM \cite{selvaraju2017grad} and attention-based saliency maps \cite{liu2023good} were largely developed for, or transferred from, recognition and natural language processing. As a result, they tend to overlook the compositional, structural, and temporal information that more robust visual reasoning needs in domains such as biomedical imaging and remote sensing \cite{stammer2021clevr,zhang2021salient,petrova2024spatiotemporal}.

The rise of large-scale vision–language models (VLMs) and multimodal large language models (MLLMs) brings both opportunities and difficulties for visual reasoning. Many recent approaches fold language, prompts, or large pretrained models into reasoning pipelines \cite{alayrac2022flamingo,li2023blip2,liu2023llava,gupta2023visual}, and prompting techniques such as multimodal chain-of-thought have begun to elicit explicit step-by-step inference \cite{zhang2023multimodalcot,wei2022chain}. However, this cross-modal integration can obscure vision-specific inference and raises the question of whether such models truly ``understand'' visual content or mainly exploit linguistic priors. Most existing surveys either predate this shift or treat individual pieces of it (e.g., VQA, scene graphs, neuro-symbolic AI, or chain-of-thought) in isolation. Few take a unified view that connects classical structured reasoning with the VLM/MLLM paradigm and that jointly considers methodological evolution, the limits of current evaluation practice, and future directions.

This survey concentrates on reasoning methods that operate primarily on visual input and proposes a unified analytical framework for visual reasoning. We classify visual reasoning into five types (relational, symbolic, temporal, causal, and commonsense) and examine each in terms of technical evolution, strengths, and limitations. Representative methods include graph neural networks, modular attention mechanisms, neuro-symbolic pipelines, and memory-augmented architectures \cite{park2020visualcomet,mohammadi2024augmented}. We also review three dimensions of evaluation (functional correctness, structural consistency, and causal robustness \cite{cui2023braingb,heydaribeni2024distributed,zhang2022predicting}) and discuss their limits in generalizability, reproducibility, and explanatory power.

The main \textbf{contributions} of this work are as follows:

\begin{itemize}
    \item \textbf{Unified taxonomy and analytical framework of visual reasoning.} We organize five types of reasoning (relational, symbolic, temporal, causal, and commonsense) along two orthogonal axes, reasoning goals and representational modes, and bind them together with methods, tasks, and evaluation metrics, clarifying the intrinsic relations, distinctions, and complementarities among reasoning paradigms.

    \item \textbf{Bridging classical and foundation-model reasoning.} Beyond graph-based, symbolic, and modular pipelines, we explicitly position the recent wave of VLM/MLLM-based reasoning, namely multimodal chain-of-thought, visual programming, tool use, and test-time reasoning, within the same taxonomy. We analyze how it reshapes each reasoning type, assess its technical strengths and scope of applicability, and identify where it still falls short of faithful, verifiable inference.

    \item \textbf{A prospective research agenda for next-generation visual reasoning systems.} We outline deficiencies in scalability, cross-domain generalization, and explainability under weak supervision common to state-of-the-art methods, and propose research directions including the combination of symbolic and subsymbolic reasoning, cross-domain adaptive architecture design, and multi-type reasoning benchmarking and testing protocols.
\end{itemize}

\begin{figure}
    \centering
    \includegraphics[width=1\linewidth]{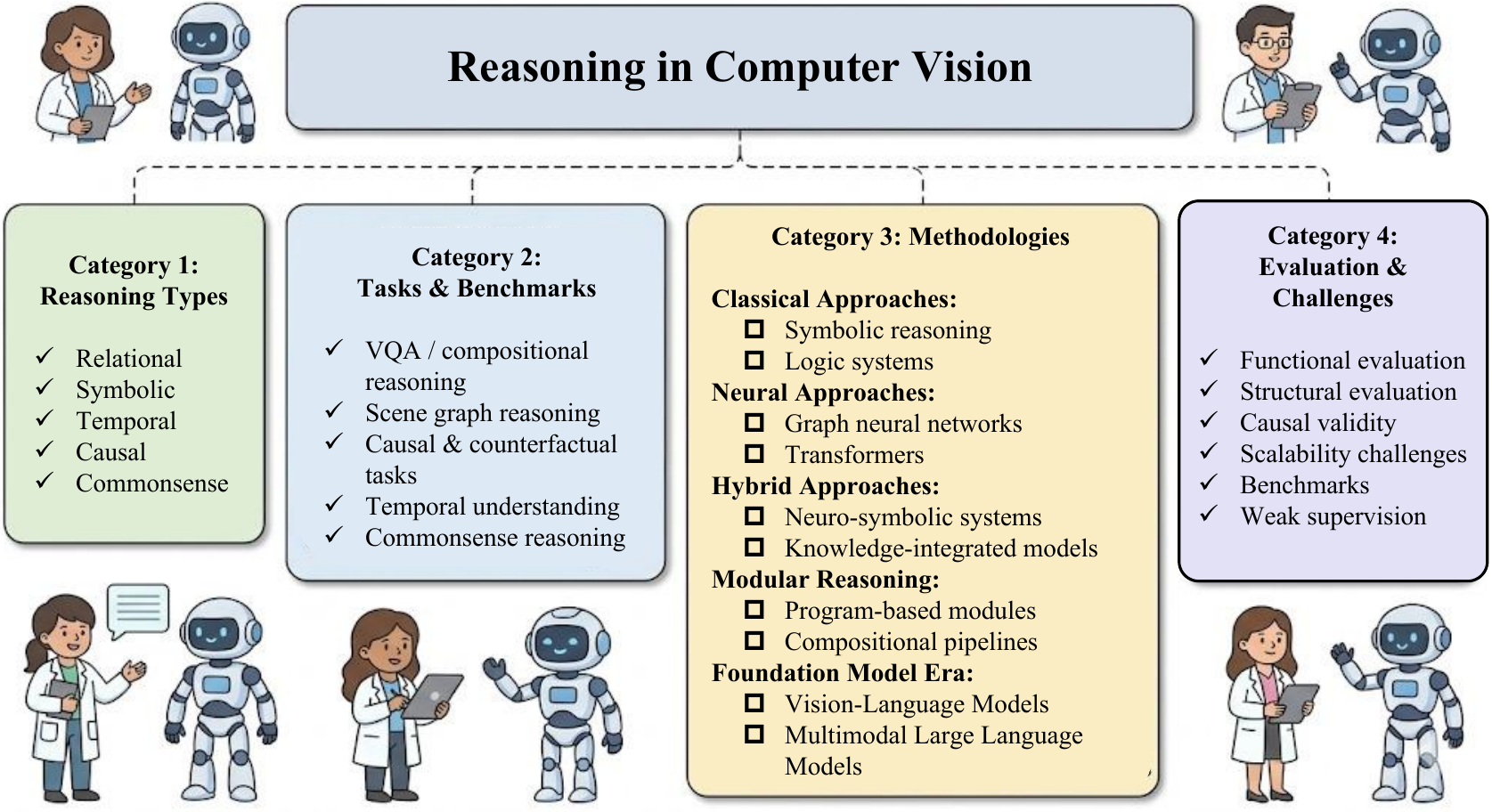}
    \caption{Structure of this survey: five reasoning types (\S3) are studied across reasoning-centric tasks and benchmarks (\S4), methodologies spanning classical to foundation-model approaches (\S5), and a three-dimensional evaluation view together with open challenges (\S6--7).}
    \label{fig:overview}
\end{figure}

To make the scope and incremental value of this survey explicit, Section~\ref{sec:scope} describes our literature-search methodology and contrasts our coverage with representative existing surveys (Table~\ref{tab:survey-comparison}). The remainder of this paper is organized as follows. Section~\ref{sec:scope} states the survey scope, search methodology, and differences from prior surveys. Section~\ref{sec:foundations} presents the theoretical foundations and taxonomy of visual reasoning. Section~\ref{sec:vision_tasks} reviews the core tasks and associated benchmark datasets. Section~\ref{sec:reasoning methods} surveys representative reasoning methodologies, spanning symbolic, relational, causal, modular, and VLM/MLLM-based models. Section~\ref{sec:evaluation} discusses evaluation protocols and their limitations. Section~\ref{sec:challenges} outlines key challenges and emerging research directions. Figure~\ref{fig:overview} illustrates the overall structure of the survey.

\section{Scope and Survey Methodology}
\label{sec:scope}

\subsection{Scope}

This survey focuses on \emph{reasoning that operates primarily on visual input}, that is, inference over object attributes, relations, dynamics, and causal structure in images and video. We deliberately include three bodies of work that are often treated separately. The first is classical structured reasoning over scene graphs, logic programs, and structural causal models. The second is neural and modular reasoning with graph neural networks, relation networks, neural module networks, and spatio-temporal transformers. The third is the recent paradigm of reasoning with vision–language models (VLMs) and multimodal large language models (MLLMs), including visual chain-of-thought, visual programming, and tool-augmented inference. We exclude work in which the image is purely incidental (e.g., document or chart parsing with no visual relational structure) and pure language reasoning without a visual grounding component. The goal is not exhaustive enumeration but a \emph{unified, comparative} account that connects these strands across reasoning types, methodologies, and evaluation protocols.

We emphasize that, in current practice, ``reasoning in computer vision'' is operationalized predominantly through tasks at the interface of vision and language, namely visual question answering and its compositional variants, scene-graph generation, video question answering, and visual commonsense and causal benchmarks. Consequently, much of the literature we review (and most public benchmarks) is VQA-centric, and we use these tasks as the empirical substrate for comparison. We treat purely perceptual problems (detection, segmentation, recognition) as the \emph{input layer} to reasoning rather than reasoning itself, and we flag throughout where conclusions drawn from VQA-style benchmarks may not transfer to broader vision tasks such as embodied perception or open-world video understanding.

\subsection{Search Methodology}

To assemble the reviewed literature, we conducted a structured search over DBLP, IEEE~Xplore, the ACM Digital Library, Scopus, and arXiv, complemented by Google Scholar for citation tracking. We queried combinations of the terms \textit{visual reasoning}, \textit{compositional / relational / causal / temporal / commonsense reasoning}, \textit{visual question answering}, \textit{scene graph}, \textit{neuro-symbolic}, \textit{visual chain-of-thought}, and \textit{vision–language model}, restricted to the period 2015--2025. Records were screened by title and abstract against the following \emph{inclusion criteria}: the work (a) proposes, analyzes, or evaluates a reasoning mechanism over visual input, (b) introduces a benchmark or evaluation protocol for such reasoning, or (c) is a survey of an adjacent area used for positioning. We \emph{excluded} pure perception/recognition papers without an inferential component, short workshop abstracts, and non-archival reports. We then applied backward and forward snowballing on highly cited works and recent foundation-model papers to capture fast-moving 2023--2025 developments. We characterize this process as a \emph{structured but narrative} literature review rather than a formal systematic review: it is designed for breadth and representativeness across reasoning types and methodologies, and we make no claim to an exhaustive, fully logged audit of every matching record. The retained corpus underpins the taxonomy, method tables, and evaluation discussion in the following sections.

\subsection{Relation to Existing Surveys}

Prior surveys tend to examine a single slice of the landscape, such as VQA with MLLMs~\cite{kuang2025vqamllmsurvey}, multimodal large language models in general~\cite{yin2024mllmsurvey}, vision–language models for recognition and transfer~\cite{zhang2024vlmsurvey}, multimodal chain-of-thought~\cite{wang2025mcotsurvey}, scene-graph generation~\cite{chang2023scenegraphsurvey}, neuro-symbolic AI~\cite{bhuyan2024neurosymbolicsurvey}, causal visual representation learning~\cite{liu2022causalvisualsurvey}, or compositional generalization~\cite{lin2023compgensurvey}. Table~\ref{tab:survey-comparison} contrasts these works along the axes that matter for our purpose, namely breadth across reasoning \emph{types}, coverage of \emph{methodologies}, treatment of \emph{evaluation protocols}, inclusion of the \emph{VLM/MLLM era}, and, most importantly, whether the survey provides a \emph{unified cross-type analysis} that relates reasoning types to one another rather than reviewing them in isolation. To our knowledge, no existing survey jointly spans the five reasoning types, the classical-to-foundation-model methodological spectrum, and the three-dimensional (functional, structural, causal) evaluation view that we develop here.

\begin{table}[!ht]
\centering
\caption{Positioning of this survey relative to representative recent surveys. \textbf{Y} denotes coverage in depth, \textbf{P} partial or secondary coverage, and \textbf{N} no coverage. ``Reasoning breadth'' counts how many of the five reasoning types (relational, symbolic, temporal, causal, commonsense) are treated. Ratings reflect the authors' assessment against these criteria; the entries for this survey are substantiated by the correspondingly indexed sections (types in \S3, methods in \S5, evaluation in \S6, VLM/MLLM coverage in \S5.6, and the cross-type analysis in \S3 and \S5.7). The final row is the authors' self-assessment rather than an independent evaluation. To avoid over-claiming, we (i) emphasize in bold only the two axes on which this survey is genuinely differentiated from prior work (reasoning breadth and unified cross-type analysis), and (ii) conservatively rate our own \emph{evaluation-protocol} contribution as partial (\textbf{P}), since Section~\ref{sec:evaluation} offers a conceptual three-dimensional (functional/structural/causal) taxonomy and a deliberately coarse coverage map rather than a fully validated empirical protocol.}
\label{tab:survey-comparison}
\footnotesize
\renewcommand{\arraystretch}{1.2}
\rowcolors{2}{rowblue}{white}% 从第2行起，浅蓝/白交替
\begin{tabularx}{\linewidth}{>{\raggedright\arraybackslash}p{3.5cm} c >{\centering\arraybackslash}X >{\centering\arraybackslash}X >{\centering\arraybackslash}X >{\centering\arraybackslash}X >{\centering\arraybackslash}X}
\toprule
\rowcolor{headerblue}
\textbf{Survey} & \textbf{Year} & \textbf{Reason. breadth} & \textbf{Methods} & \textbf{Eval. protocols} & \textbf{VLM/ MLLM} & \textbf{Unified cross-type} \\
\midrule
VQA with MLLMs~\cite{kuang2025vqamllmsurvey}        & 2025 & 2/5 & P & P & Y & N \\
MLLMs (general)~\cite{yin2024mllmsurvey}            & 2024 & 2/5 & Y & P & Y & N \\
VLMs for vision~\cite{zhang2024vlmsurvey}          & 2024 & 1/5 & Y & P & Y & N \\
Multimodal CoT~\cite{wang2025mcotsurvey}            & 2025 & 3/5 & P & P & Y & P \\
Scene graphs~\cite{chang2023scenegraphsurvey}      & 2023 & 1/5 & Y & P & N & N \\
Neuro-symbolic AI~\cite{bhuyan2024neurosymbolicsurvey} & 2024 & 2/5 & Y & P & P & N \\
Causal visual repr.~\cite{liu2022causalvisualsurvey} & 2022 & 1/5 & P & P & N & N \\
Compositional gen.~\cite{lin2023compgensurvey}     & 2023 & 2/5 & P & P & N & P \\
\rowcolor{headerblue}
\textbf{This survey}                               & 2025 & \textbf{5/5} & Y & P & Y & \textbf{Y} \\
\bottomrule
\end{tabularx}
\end{table}

\section{Foundations}
\label{sec:foundations}

\subsection{Definition}

Before listing categories, we make an organizing distinction that is often left implicit in the literature and that we maintain throughout the survey. Visual reasoning can be characterized along two largely \emph{orthogonal} axes. The first is the \emph{reasoning goal}, that is, what kind of inference is required, such as relating entities, ordering events in time, attributing cause, or invoking background and commonsense knowledge. The second is the \emph{representational or computational mode}, that is, how that inference is realized, whether by sub-symbolic neural computation, explicit symbolic and logical manipulation, neuro-symbolic combinations, or probabilistic graphical models. These axes are not interchangeable. Relational, temporal, causal, and commonsense reasoning can each be expressed symbolically (e.g., temporal logic, causal logic programs, commonsense rule bases) \emph{or} sub-symbolically, and causality in particular can be captured by structural causal models as well as by Bayesian and probabilistic graphical models~\cite{pearl1988probabilistic, koller2009pgm, garcez2023neurosymbolic3rdwave}. Accordingly, what the visual-reasoning literature commonly labels ``symbolic reasoning'' is better understood as a representational mode than as a reasoning goal of its own.

For continuity with established terminology, we nonetheless organize the discussion around the five categories most frequently used in the field, namely symbolic, relational, causal, temporal, and commonsense (intent-driven) reasoning~\cite{mu2023learning, rajani2019explain, meng2023mass, she2023symbolic, hessel2022multi,gupta2023visual,thawakar2025llamav,chen2024visual,tan2025reason}. We read ``symbolic'' as the representation axis and the remaining four as reasoning goals, and we make this mapping explicit wherever it affects the analysis (e.g., the method taxonomy in Section~\ref{sec:reasoning methods} and the representations in Section~\ref{sec:foundations}).

\textbf{Symbolic reasoning} (a representational mode rather than a content domain) involves the manipulation of discrete structured abstractions derived from visual input. It typically relies on the transformation of raw imagery into intermediate symbolic representations such as scene graphs, logic trees, or programs, followed by rule-based inference, and it can serve as the substrate for the other four reasoning goals (e.g., temporal logic for ordering, causal logic for intervention, commonsense rule bases for intent). This yields highly interpretable reasoning pipelines. For example, a symbolic model might deduce: "if object A is to the left of object B and B is red, then A is to the left of a red object."  Although this explicit traceability is advantageous for verification and explainability, the method requires annotated semantic information, such as segmentation masks or object attributes, which are often non-trivial to extract directly from pixels~\cite{ahmed2023neuro-symbolic,cheng2025neural,li2024neuro,shindo2024learning}.

\textbf{Relational reasoning}, in contrast, emphasizes the discovery and modeling of pairwise or higher-order relations among visual entities. These relations can span spatial (for example, “above,” “behind”), functional (for example, “holding,” “supporting”), or semantic (for example, “part of”) dimensions. Rather than using predefined rules, relational models, often implemented with graph neural networks (GNNs), transformers, or relation networks, learn to capture dependencies implicitly through architecture design. Such capabilities are crucial for tasks such as scene graph generation, visual reference expression comprehension, and analogy-based image reasoning~\cite{chang2022reasoning, xu2022scene,kim2024higher,wang2024earthvqa,liu2024chronobridge,ahmed2024analogical}.

\textbf{Causal reasoning} extends beyond correlation by modeling mechanisms through which one event influences another. In visual domains, this often involves identifying directional dependencies (e.g., “the glass shattered because it fell”) rather than mere co-occurrence. Methods such as Structural Causal Models (SCMs), counterfactual estimators, and do-calculus provide a foundation for deriving interventional and hypothetical conclusions from visual data. These approaches are particularly relevant in video-based reasoning, where event chronology encodes rich causal signals~\cite{zhang2023causalvideo, brehmer2023causal,chi2024unveiling,zhu2024deep,gupta2025causal}.

\textbf{Temporal reasoning} addresses dynamic visual content that unfolds over time. Tasks such as motion prediction, action segmentation, or event ordering require understanding both the temporal evolution of individual objects and the interactions between entities. Depending on the task complexity, architectures range from recurrent neural networks like LSTMs and temporal convolutional networks (TCNs) to attention-based spatio-temporal transformers that capture long-range dependencies. Temporal reasoning is especially vital in surveillance, video quality assurance (QA), and autonomous perception~\cite{zhang2024sea, li2023temporal, cao2023motionreasoner,xiong2024large,yuan2024back,fatemi2024test}.

\textbf{Commonsense and intent reasoning} concern the implicit understanding of human actions, goals, and contextual cues. It enables vision systems to make plausible inferences even in ambiguous or partially observable settings. For example, interpreting that a person reaching for a cup intends to drink from it involves a combination of physical context, social norms, and prior knowledge. This class of reasoning plays a central role in real-world applications, including assistive robotics, driver intention prediction, and social scene analysis. Models typically incorporate external knowledge bases, memory-augmented networks, or pre-trained language models to support this form of reasoning~\cite{qin2023knowledgevision, chao2022intentnet, lin2023raven,wang2024commonsensevis}.

\begin{figure}
    \centering
    \includegraphics[width=1\linewidth]{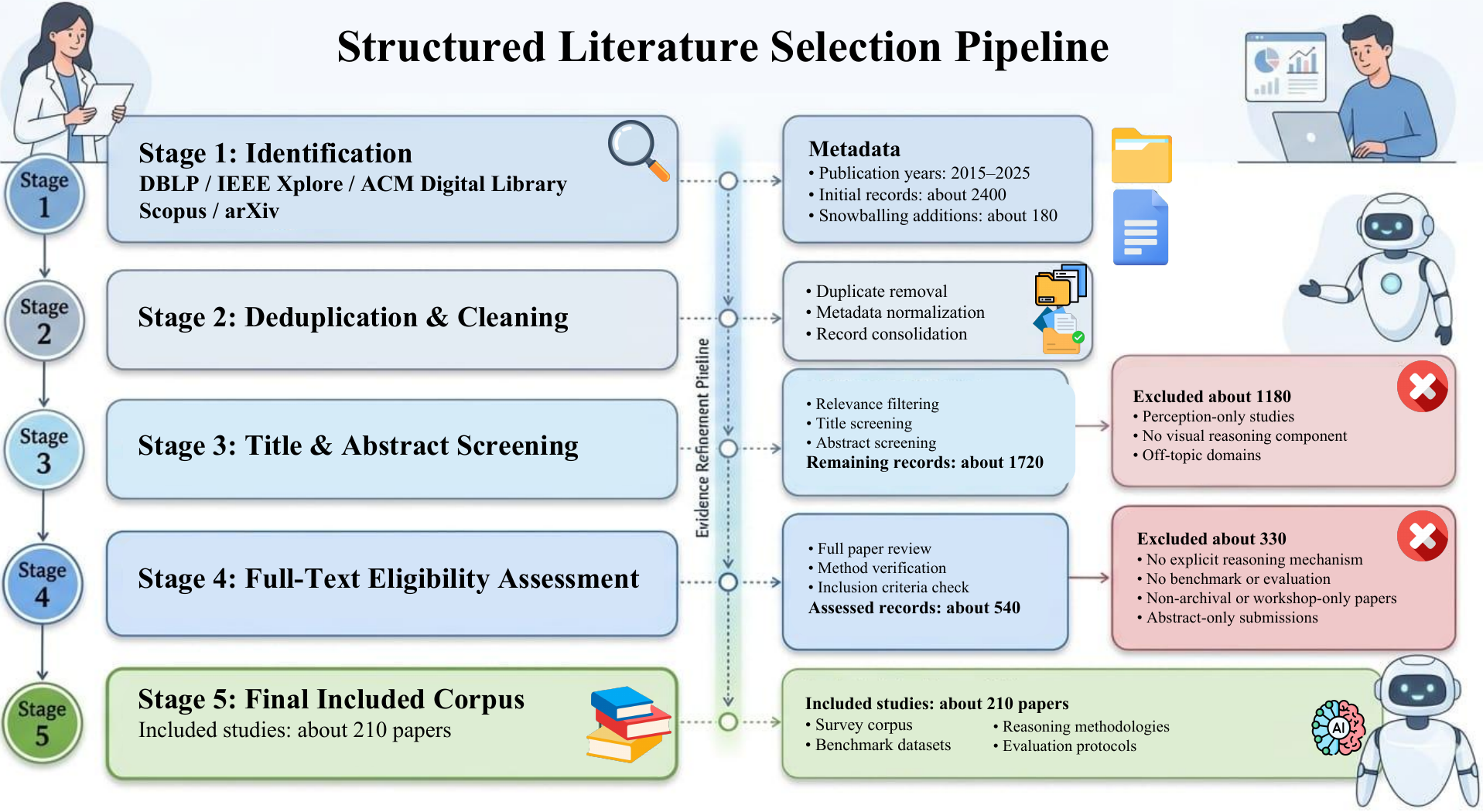}
    \caption{The five reasoning types as an interconnected continuum. Each node lists its dominant method family, and edge labels denote the conceptual bridges through which adjacent reasoning types compose. Commonsense reasoning is increasingly realized through vision–language and multimodal large language models (Section~\ref{subsec:vlm-reasoning}).}
    \label{fig:taxonomy-relations}
\end{figure}

These five types are not mutually exclusive. In practice they form a connected continuum, and most real-world tasks recruit several at once. Figure~\ref{fig:taxonomy-relations} summarizes the taxonomy together with the dominant method family for each type and the conceptual ``bridges'' that link adjacent types. For example, symbolic abstractions provide the structured relations that relational models operate over, temporal order supplies the raw signal that causal reasoning converts into cause–effect chains, and commonsense reasoning integrates situational context with inferred intent. Viewing the types as an interrelated whole, rather than as isolated silos, is the organizing principle of this survey and motivates the cross-type analysis in the sections that follow.

\definecolor{catTypes}{RGB}{212,230,214} % 柔和绿  —— Relational
\definecolor{catTasks}{RGB}{208,224,240} % 柔和蓝  —— Symbolic
\definecolor{catMeth}{RGB}{244,232,208}  % 柔和琥珀 —— Temporal
\definecolor{catEval}{RGB}{234,216,224}  % 柔和藕粉 —— Causal
\definecolor{catCmn}{RGB}{224,219,236}   % 柔和雪青 —— Commonsense
\definecolor{rootgray}{RGB}{216,222,230} % 冷灰蓝  —— 中心节点

\subsection{Representations}

To support the aforementioned reasoning types, vision systems rely on rigorous mathematical formalisms and data representations. These foundations not only encode visual semantics but also facilitate structured and interpretable inference. In the following, we summarize four core representational approaches used across visual reasoning models.

\textbf{Scene Graphs.} Scene graphs are structured representations of visual content, where nodes denote objects and edges represent their relationships. Formally, a scene graph is a directed graph \( G = (V, E) \), where \( V \) represents detected entities and \( E \subseteq V \times R \times V \) defines labeled relationships, with \( R \) capturing types such as “holding” or “next to.” These graphs are fundamental for relational reasoning and are widely used in tasks such as VQA, captioning, and visual grounding~\cite{yang2018graph, xu2017scene, tang2020unbiased,gao2024graphdreamer,li2024pixels,epstein2024graphbased}.

\textbf{Structural Causal Models (SCMs).} SCMs provide a principled framework for modeling causality in visual data. They represent each variable \( X_i \) as a function \( f_i \) of its causal parents \( \text{PA}_i \) and some exogenous noise \( U_i \). The causal graph \( \mathcal{G} \) defines the dependencies between variables. Using the do-calculus, we can estimate intervention outcomes with expressions such as \( P(Y|do(X=x)) \). SCMs are especially valuable for video scenarios that involve interventions, allowing counterfactual analyses such as "What would have happened if object A did not move?" ~\cite{pearl2009causality, zhang2020causal, schoelkopf2021towards,ashwani2024cause,fujii2025estimating}.

\textbf{Logic-Based Programs.} Logic-driven reasoning employs formal rule sets on symbolic visual abstractions. Using first-order logic, these systems evaluate statements such as:
\begin{equation}
\forall x, y \; ( \text{OnTop}(x, y) \land \text{Red}(x) \Rightarrow \text{AboveRedObject}(y) )
\end{equation}
These logic rules are interpretable and amenable to integration with neural modules, especially in compositional or programmatic tasks such as symbolic VQA or explanation of a structured scene~\cite{lampinen2023complementary, han2020explaining, evans2023symbolic,shindo2024learning}.

\textbf{Graph and Attention Mechanisms.} Graph Neural Networks (GNNs) and attention models serve as the computational backbone of many modern visual reasoning. In GNNs, each node \( v \) updates its representation based on its neighbors \( \mathcal{N}(v) \) using the following:
\begin{equation}
h_v^{(k)} = \sigma \left( \sum_{u \in \mathcal{N}(v)} W^{(k)} h_u^{(k-1)} + b^{(k)} \right)
\end{equation}
These models support structured message passing and are used in scene graph reasoning, spatial inference, and knowledge-aware tasks~\cite{kipf2016semi,besta2024parallel,chen2023graph,jin2023gnnlens,lee2024privacy,yin2024dynamic,yoon2023unbiased,zheng2023temporal}.

Transformer-based attention, meanwhile, computes relevance between entities via:
\begin{equation}
\text{Attention}(Q, K, V) = \text{softmax} \left( \frac{QK^\top}{\sqrt{d_k}} \right) V
\end{equation}
These mechanisms are indispensable in capturing long-range dependencies, particularly in temporal and multimodal contexts~\cite{vaswani2017attention, dosovitskiy2020image, parmar2018image,kumar2024spatio,zhang2024s2wat,zheng2024spatio}.

Each of these representational tools introduces trade-offs between expressivity, scalability, and interpretability. Their selection profoundly influences the reasoning capabilities of vision systems and the transparency of their outputs. As such, they form the conceptual and operational bedrock upon which explainable visual reasoning is built.

\section{Reasoning-Centric Tasks}
\label{sec:vision_tasks}

As summarized in Table~\ref{tab:vision_tasks}, these reasoning-centric tasks fall into five major types with distinct capabilities and benchmarks. Visual reasoning moves computer vision from passive perception toward active understanding, where models interpret structured relationships, model temporal dynamics, and infer causal mechanisms. These capabilities connect low-level visual features with higher-level inference. This section presents a taxonomy of core reasoning-centric tasks that demand more than pixel-level recognition. For each category, we describe its reasoning objectives, representative datasets, and typical evaluation settings, which set up the systematic comparison and methodological analysis in later sections.

\begin{table}[!ht]
\centering
\caption{Comparison of core visual reasoning task types, their primary reasoning capabilities, and representative datasets. Task descriptions are consistent with Sec.~\ref{sec:vision_tasks}. Note: While these datasets are widely used benchmarks, many rely heavily on synthetic scenes or curated domains, limiting their direct transferability to unconstrained real-world settings.}
\footnotesize
\label{tab:vision_tasks}
\rowcolors{2}{rowblue}{white}% 从第2行起，浅蓝/白交替
\begin{tabular}{>{\raggedright\arraybackslash}m{3cm} >{\raggedright\arraybackslash}m{8cm} >{\raggedright\arraybackslash}m{4cm}}
\toprule
\rowcolor{headerblue}
\textbf{Task Type} & \textbf{Reasoning Capability} & \textbf{Representative Datasets} \\
\midrule
\textbf{Relational Reasoning} & Modeling spatial, semantic, and functional relationships among objects in a scene. & Visual Genome~\cite{krishna2017visual}, VRD~\cite{lu2016visual}, Open Images V6~\cite{kuznetsova2020open} \\
\textbf{Compositional VQA} & Performing multi-step symbolic, logical, or compositional reasoning to answer structured visual questions. & CLEVR~\cite{johnson2017clevr}, GQA~\cite{hudson2019gqa}, VQA v2~\cite{goyal2017vqa} \\
\textbf{Commonsense and Intent Reasoning} & Inferring latent goals, motivations, and physical affordances from contextual and background knowledge. & VCR~\cite{zellers2019recognition}, VisualCOMET~\cite{park2020visualcomet}, Social IQa~\cite{sap2019social} \\
\textbf{Causal and Counterfactual Reasoning} & Identifying cause–effect dynamics and hypothesizing alternative outcomes under hypothetical interventions. & CausalVQA~\cite{yang2022causalvqa}, ImageNet-9 (Backgrounds Challenge)~\cite{lopez2023imagenetcausal}, SCM Toolkit~\cite{pearl2009causality} \\
\textbf{Temporal Reasoning} & Understanding visual event sequences, temporal dependencies, and action transitions over time. & SSv2~\cite{goyal2017something}, Charades~\cite{sigurdsson2016hollywood}, TVQA~\cite{lei2018tvqa} \\
\bottomrule
\end{tabular}
\end{table}

\subsection{Relational Reasoning}

Relational reasoning assesses a model’s ability to identify and interpret interactions among visual entities, encompassing spatial relations (e.g., “above,” “under”), functional dependencies (e.g., “holding,” “riding”), and semantic hierarchies (e.g., “part-of,” “belongs-to”). Two core tasks dominate this domain: visual relationship detection (VRD) and scene graph generation (SGG).

In VRD, the goal is to extract structured relational triplets such as “man riding bicycle” or “dog under table.” SGG extends this by generating a comprehensive graph representation that encodes all detected objects and their pairwise relationships. Recent advances employing graph neural networks (GNNs), attention-based mechanisms, and transformer architectures have achieved state-of-the-art performance~\cite{xu2017scene, zellers2018neural, tang2020unbiased}. Benchmark datasets such as Visual Genome~\cite{krishna2017visual}, VRD~\cite{lu2016visual}, and Open Images V6~\cite{kuznetsova2020open} are widely adopted. Importantly, the structured outputs from these tasks support a range of downstream applications, including visual question answering, image captioning, and embodied AI navigation. 

Nonetheless, current relational reasoning benchmarks suffer from significant long-tail biases and limited relation diversity, leading to models that excel on frequent relation types but fail to generalize to rare or context-dependent interactions. Moreover, most datasets lack fine-grained annotations for functional or temporal dependencies, restricting progress toward robust, real-world relational understanding~\cite{tang2020unbiased, chang2023scenegraphsurvey}.

\subsection{Compositional Visual Question Answering (VQA)}

Compositional VQA extends standard visual question answering by requiring multi-step reasoning over an image. Instead of retrieving direct answers, models must sequentially perform operations such as filtering, counting, comparison, and logical deduction. Benchmark datasets like CLEVR~\cite{johnson2017clevr} and GQA~\cite{hudson2019gqa} are explicitly designed to test such capabilities, using structured scenes and synthetic questions to reduce the impact of dataset biases and shortcut learning.

For example, the query “Are there more red cubes than blue spheres?” demands a chain of reasoning, in which the model must detect relevant objects, classify their colors and shapes, count each category, and compare the counts. State-of-the-art approaches often employ neural module networks, symbolic reasoning pipelines, or hybrid neuro-symbolic models~\cite{andreas2016neural, hu2017learning}, producing intermediate outputs, such as executable programs or visual traces, that enhance interpretability and facilitate debugging~\cite{zhang2022surveyxai, abbas2025crossmodality}. These properties make compositional VQA particularly relevant in domains where explainability, step-by-step verification, and robust generalization are critical.  

Yet, the controlled nature of current benchmarks imposes a ceiling on real-world applicability. Scene diversity and linguistic richness remain limited, with synthetic environments failing to capture the ambiguity, occlusions, and background noise present in natural images. As a consequence, models that excel in benchmark settings often experience significant performance drops in unconstrained scenarios, highlighting the gap between experimental evaluations and deployment-level robustness~\cite{hudson2019gqa, kuang2025vqamllmsurvey}.

\subsection{Commonsense and Intent Reasoning}

Some reasoning problems require inference beyond explicit visual signals, drawing instead on contextual knowledge, intuitive physics, or social conventions. Tasks in this category include predicting human intent, understanding plausible future events, and interpreting behaviors that align with real-world expectations. A classic example is inferring that a person reaching toward a glass likely intends to drink.

Datasets such as VisualCOMET~\cite{park2020visualcomet}, VCR~\cite{zellers2019recognition}, and Social IQa~\cite{sap2019social} test these capabilities by prompting models to hypothesize motivations, causal explanations or social responses. Given their under-specified nature, such tasks often require integrating language models or external knowledge graphs~\cite{luo2024survey}. These types of reasoning are particularly relevant for interactive agents, socially aware systems, and assistive technologies operating in dynamic human-centric settings.

Despite these advances, most existing benchmarks remain constrained by static imagery and brief textual prompts, which inadequately capture the fluidity of intentions, the subtleties of cultural norms, and the multi-agent interactions typical of real-world contexts. As a result, models that excel in curated testbeds may struggle to generalize when faced with ambiguous, evolving, or culturally nuanced scenarios, underscoring the need for richer, longitudinal, and culturally diverse evaluation settings~\cite{luo2024survey, park2020visualcomet}.

\subsection{Causal and Counterfactual Reasoning}

Causal reasoning focuses on uncovering why events occur and how they might change under alternate conditions. Unlike correlation-based approaches, causal models aim to identify and reason about interventions and their consequences. Tasks in this category often include questions like “What caused the glass to fall?” or “What would happen if the object were removed?”

To support this, datasets such as CausalVQA~\cite{yang2022causalvqa}, the Backgrounds Challenge (ImageNet-9)~\cite{lopez2023imagenetcausal}, and counterfactual explanation frameworks~\cite{ivanova2024counterfactual} provide annotated visual scenarios with cause-effect chains. Formal frameworks such as Structural Causal Models (SCMs)~\cite{pearl2009causality} and counterfactual logic help encode these relations. Such reasoning is essential in domains where understanding mechanisms, not just outcomes, is critical. This includes applications in autonomous driving, diagnostics, and safety-critical decision making, where evaluating alternate futures or interventions could mitigate risks.

However, prevailing benchmarks often simplify causality into discrete and visually salient events, neglecting latent confounders, long-range dependencies, and multi-causal interactions common in real-world settings. As a result, models may excel on such datasets yet fail to capture the nuanced causal structures required for robust generalization in dynamic or safety-critical contexts~\cite{liu2022causalvisualsurvey}.

\subsection{Temporal Reasoning}

Temporal reasoning involves modeling how visual events unfold over time. Rather than interpreting isolated images, models must understand sequences: the order, duration, and transition of actions. The core tasks in this area include action recognition, event sequencing, anticipation, and temporal grounding of natural language to video content.

Datasets like Something-Something V2~\cite{goyal2017something}, Charades~\cite{sigurdsson2016hollywood}, and TVQA~\cite{lei2018tvqa}, along with recent benchmarks such as V-STaR~\cite{hu2025vstar} for spatio-temporal reasoning and TUMTraf VideoQA~\cite{chen2025tumtraf} for traffic-scene VideoQA, challenge models with tasks that rely on fine-grained motion cues and timing. Differentiating between actions such as “pretending to push” and “actually pushing” depends on subtle temporal dynamics. To model such behaviors, researchers have employed recurrent networks (e.g. LSTM), 3D convolutional architectures, and more recently spatiotemporal transformers~\cite{bertasius2021space, carreira2017quo,feng2024efficient,khan2022transformers, muller2025unsupervised}.

While these advances have substantially improved short- to mid-range temporal modeling, long-horizon prediction remains a major bottleneck. Occlusions, ambiguous transitions, and sparse observations often lead to brittle representations that fail to generalize. Overcoming these limitations may require memory-augmented architectures, richer multimodal temporal cues, and reasoning frameworks grounded in structured commonsense timelines.

\section{Methodologies}
\label{sec:reasoning methods}

This section concerns \emph{methodologies}, the computational approaches that operationalize reasoning. It builds on the representations introduced in Section~\ref{sec:foundations} (scene graphs, SCMs, logic programs, graphs and attention) and corresponds to the ``Methodologies'' block of Figure~\ref{fig:overview}, and it is distinct from the \emph{evaluation dimensions} of Section~\ref{sec:evaluation}. Consistent with the two-axis view of Section~\ref{sec:foundations}, these methodologies are organized primarily by representational/computational mode rather than by reasoning goal, since a single method family (e.g., neuro-symbolic pipelines) is routinely applied across relational, temporal, causal, and commonsense goals.

Reasoning-focused vision tasks need models to go beyond passive recognition to active inference, compositional integration, and structured knowledge modeling. Currently proposed approaches are often categorized into four prevailing inference paradigms, namely symbolic, neural, hybrid, and modular. We additionally treat a fifth, cross-cutting paradigm that has reshaped the field since 2022, namely reasoning built on top of pretrained vision–language and multimodal large language models (Section~\ref{subsec:vlm-reasoning}), which does not replace the others so much as provide a general substrate on which they are re-expressed through prompting, tool use, and test-time computation.
Symbolic approaches offer formality and verifiability but need highly structured input. Neural models adapt well to noisy and unstructured data but are largely uninterpretable. Hybrid models try to combine the strengths of both, though they face problems with representation alignment and computational cost. Modular systems support sequential, compositional reasoning, but they scale poorly in open-world scenarios.

These paradigms are not mutually exclusive, recent research increasingly focuses on multi-paradigm integration, such as embedding causal reasoning into graph-based models, incorporating differentiable logic into transformer architectures, and enhancing commonsense grounding through large-scale vision–language models.
Such trends reflect a shift toward systems that balance symbolic auditability, neural adaptability, and modular interpretability under unified frameworks, while addressing core challenges of scalability, generalization, and cross-domain robustness.

Table~\ref{tab:xai_cv_reasoning} tabulates representative models, and their architectural paradigms, reasoning scopes, and domains of applications are then discussed, followed by a comparative study of their technical strengths and enduring bottlenecks, which suggests promising future directions of visual reasoning systems.
Symbolic approaches prevail in applications involving formal reasoning traces and explicit checking of constraints, but have high annotation and preprocessing expenses. Neural methods, particularly graph- and transformer-style networks, have extensive representational capability and domain adaptability, but have difficulties with reasoning explicability and vulnerability to spurious correlations. Hybrid approaches, such as neuro-symbolic pipelines and integrated-knowledge models, have shown promise in marrying explicability and adaptability, but suffer yet from seamless integration mechanisms and efficient training schedules. Modular reasoning frameworks are characterized by compositional generalization and explainable execution, but tend to rely on carefully curated supervision and suffer scaling to realistic, complex-world scenarios.

Recent developments show that performance gains increasingly come from cross-paradigm hybrids. For example, fusing causal reasoning with graph attention networks or embedding commonsense knowledge bases into multimodal transformers shows clear promise. This convergence suggests that future visual reasoning systems will likely be multi-paradigm by design, optimized for explainability, scalability, and cross-domain robustness.

\begin{table}[H]
\centering
\footnotesize
\caption{Representative visual reasoning methodologies in computer vision, organized by reasoning type, architectural paradigm, backbone model, reasoning scope, and application domain.}
\label{tab:xai_cv_reasoning}
\setlength{\tabcolsep}{3pt}
\rowcolors{2}{rowblue}{white}% 从第2行起，浅蓝/白交替
\resizebox{\textwidth}{!}{
\begin{tabular}{@{}c p{4cm} p{3cm} p{3cm} p{1.4cm} p{4cm}@{}}
\toprule
\rowcolor{headerblue}
\textbf{Ref.} & \textbf{Method Type} & \textbf{Architecture} & \textbf{Model/Backbone} & \textbf{Scope} & \textbf{Application} \\
\midrule
\cite{mao2019neuro} & Neuro-Symbolic & Hybrid Pipeline & NS-CL & Global & Visual QA, program execution \\
\cite{yi2018neural} & Neuro-Symbolic Reasoning & CNN + Symbolic Executor & CLEVR-based & Global & Scene reasoning, logic tracing \\
\cite{she2023symbolic} & Symbolic Representations & Probabilistic Program & BPL & Global & Symbolic concept learning \\
\cite{hudson2018compositional} & Modular Reasoning & Controller-Memory & MACNet & Local & Multi-hop VQA, CLEVR \\
\cite{hu2017learning} & Neural Module Networks & Modular Neural Network & NMN-VQA & Local & Neural compositional QA \\
\cite{liu2025visionpathways} & Modular Reasoning & Modular Architecture & Stack-NMN & Global & Explainable modular VQA \\
\cite{yang2022causalvqa} & Causal Reasoning & SCM-based & CausalVQA & Local & Counterfactual visual QA \\
\cite{bengio2019meta} & Causal Disentanglement & SCM-based & Meta-Transfer Objective & Global & Learning robust causal representations \\
\cite{chang2022reasoning} & Relational Reasoning & Graph Networks & Graph Net (GN) & Global & Relational inductive biases \\
\cite{xu2022scene} & Scene Graph Reasoning & GNN & VCTree & Global & Relationship detection \\
\cite{singh2024explaingraph} & Graph-based Explanations & GNN & GNNExplainer & Local & Graph model explanation \\
\cite{jiang2019stm} & Spatio-Temporal Modeling & 3D CNN & I3D, C3D & Local & Video reasoning, action recognition \\
\cite{bertasius2021space} & Spatio-Temporal Transformers & Transformer & TimeSformer & Global & Space-time reasoning in videos \\
\cite{kao2022videoexplain} & Temporal Saliency Visualizations & RNN & Excitation BP & Local & Spatiotemporal video saliency \\
\cite{park2020visualcomet} & Commonsense Reasoning & Knowledge-Integrated & VLMs & Global & Social prediction from static images \\
\cite{zellers2019recognition} & Visual Commonsense Reasoning & Multi-modal Networks & VCR model & Global & Social interaction inference \\
\cite{qin2023knowledgevision} & Knowledge-Enhanced Reasoning & Knowledge-Integrated & OK-VQA & Global & External-knowledge VQA \\
\cite{singh2025knowledgexai} & Knowledge Integration & Fact-based & FVQA & Global & Fact-based visual reasoning \\
\cite{hu2024ruleaug} & Neuro-Symbolic Reasoning & Hybrid & Calibrated NS-VR & Global & Disentangled visual reasoning \\
\cite{robinson2025visuallogic} & Probabilistic Neuro-Symbolic & Hybrid & Prob-NMN & Global & Interpretable visual QA \\
\cite{perez2018film} & Attention-based Conditioning & CNN + Language Module & FiLMNet & Local & Feature modulation via language \\
\bottomrule
\end{tabular}
}
\end{table}

\subsection{Symbolic and Neuro-Symbolic}

Symbolic approaches rely on discrete, rule-based inference applied to structured visual representations, such as logic programs, scene graphs, or grammars. Their primary strength lies in transparency, as each reasoning step can be explicitly inspected and validated. This makes them particularly well-suited for domains requiring rigorous auditability and formal guarantees. However, symbolic methods typically demand clean, well-structured inputs and often fail to cope with the ambiguity and noise present in real-world visual data, limiting their scalability and robustness.

\begin{figure}
    \centering
    \includegraphics[width=1\linewidth]{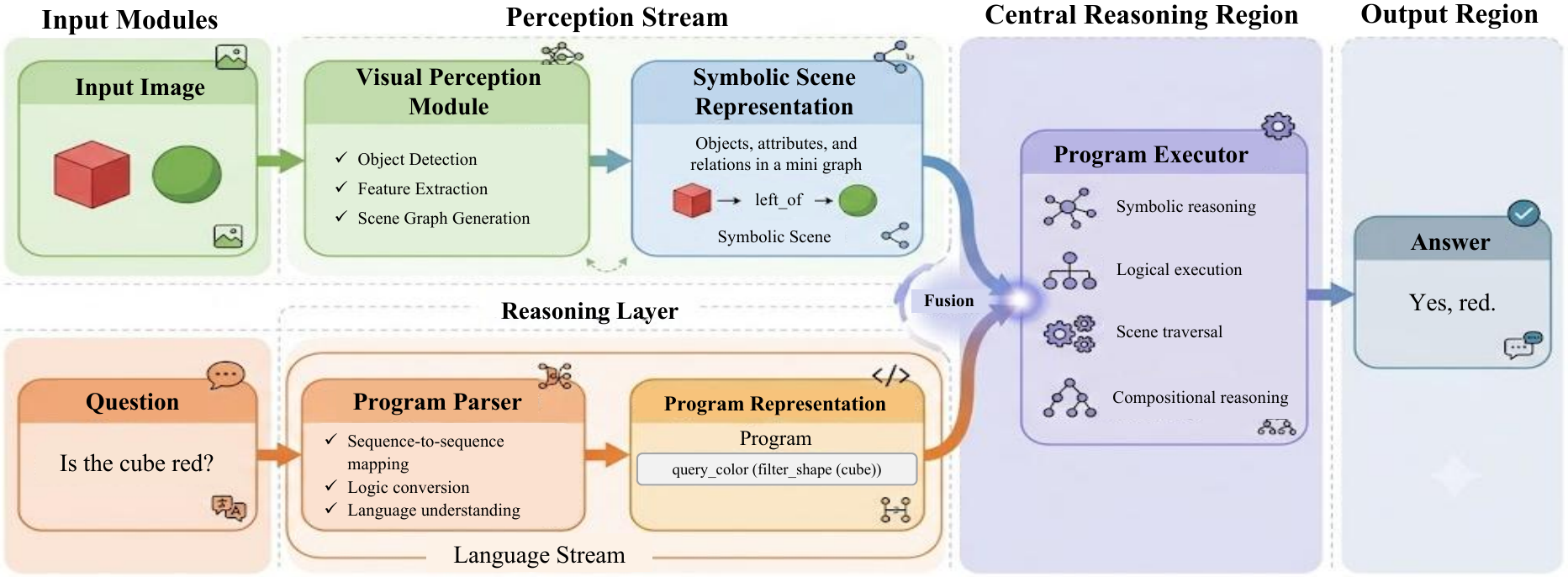}
    \caption{Neuro-symbolic reasoning pipeline. A visual stream parses the image into a symbolic scene representation, while a language stream parses the question into an executable program. A symbolic program executor then runs the program over the scene to produce a verifiable answer.}
    \label{fig:neuro-symbolic}
\end{figure}

In order to make up these deficiencies, neuro-symbolic models integrate neural perception and symbolic inference \cite{li2023variational,lake2023human,lin2025fuzzy}. As shown in Figure~\ref{fig:neuro-symbolic}, a basic pipeline deploys an object detection or scene graph generation module to convert unstructured images into symbolic objects and a program parser to convert natural-language queries into formal logic programs. A program executor then does symbolic reasoning across these structured representations to produce explicit reasoning traces and output answers. An exemplary instance of such an approach is the Neuro-Symbolic Concept Learner (NS-CL) \cite{mao2019neuro}, which outlines this two-stage approach by encoding images as symbolic objects through convolutional networks and then applying symbolic execution to support reasoning at the high level. These hybrid models have been shown to excel on structured data like CLEVR \cite{johnson2017clevr}, reaching remarkable compositional generalization and interpretability.

These approaches are most useful in applications of logical chaining, concept grounding, and explanation generation, where intermediate reasoning traces are critical. However, their reliance on carefully curated datasets and hand-designed symbolic vocabularies persists as an obstacle to deployment in open-world scenarios. Increasingly, future work investigates learning of symbolic abstractions end-to-end from unstructured visual observations, including physics-constrained symbolic regression from imagery~\cite{yu2025physics_symreg}, so that neuro-symbolic systems may close the gap between formal reasoning and the perceptual abundance of everyday imagery.

The two-stage ``parse-then-execute'' pipeline above is only one point in a broader design space. A complementary family of \emph{end-to-end} neuro-symbolic systems makes the symbolic layer differentiable so that perception and reasoning are trained jointly: DeepProbLog couples neural predicates with probabilistic logic programming~\cite{manhaeve2018deepproblog}, Logic Tensor Networks ground first-order logic in real-valued tensors for gradient-based satisfaction~\cite{badreddine2022ltn}, and Logical Neural Networks embed weighted logical constraints directly into network structure~\cite{riegel2020lnn}. Orthogonally, rather than assuming a fixed rule set, \emph{inductive} approaches learn the rules themselves: differentiable inductive logic programming recovers explanatory rules from noisy data~\cite{evans2018dilp}, connecting modern systems to the long-standing inductive logic programming tradition~\cite{cropper2022ilp30}. Recent surveys document the rapid evolution and open challenges of this area, including the move away from post-hoc analysis toward integrated learning and reasoning~\cite{garcez2023neurosymbolic3rdwave, nawaz2025neurosymbolicreview, khan2025nsvisualreasoning}.

Within vision specifically, recent advances explore more flexible pipelines, including differentiable logic solvers that enable gradient-based optimization \cite{marra2024statistical}, and neural-symbolic generative models that produce symbolic programs as latent structures \cite{li2024logicity}. Other work leverages large vision–language models (VLMs) to generate symbolic scene graphs or logical forms from raw images \cite{li2024pixels,chen2024scene}, thereby reducing reliance on manually curated vocabularies. However, challenges remain in symbol grounding under severe domain shifts, aligning symbolic spaces learned from vision and language, and scaling reasoning to unbounded open-world ontologies. A promising research direction is to integrate causal discovery with symbolic abstractions, enabling systems to reason not only about what is observed but also about why it occurs.

\subsection{Relational and Graph-Based Neural Reasoning}

Understanding relationships across objects is the root of visual reasoning \cite{jiao2023graph,zhou2024graph}. Neural graph models make such relationships explicit via object-to-node and interaction-to-edge mapping. Common methods are Graph Neural Networks (GNNs), Graph Attention Networks (GATs), and Relation Networks (RNs)~\cite{scarselli2009graph, santoro2017simple} that enable relational inference via contextual information propagation through graph representations.

As shown in Figure~\ref{fig:scene-graph-reasoning}, a scene is modeled as a graph where each detected object (e.g., Obj A–D) is a node, and edges (rel1–rel4) indicate pairwise relationships like spatial nearness, semantic co-occurrence, or physical interaction. Relational reasoning unfolds through message passing, where node features are repeatedly refined through aggregating information from neighbors. For instance, Obj~A can affect reasoning about Obj~D through both A$\rightarrow$B$\rightarrow$D and A$\rightarrow$C$\rightarrow$D, allowing for higher-order multi-hop inference. Explicit relational structure facilitates tasks like scene graph completion, object interaction recognition, and multi-hop VQA, and reveals interpretable in-between states.

Even though highly successful, however, such methods also have intrinsic limitations. The performance heavily relies on object detection and relation annotation precision, which is prone to be put through cascading errors once deployed to real-world tasks. Scaling to sparsely populated scenes or highly dynamic scenes is also problematic due to the computational complexity of propagating messages through large graphs. More recent works try to release these limitations through multimodal relational cue integration, usage of pretrained vision–language models as relation priors, and sparsity-awareness or hierarchical graph representations that encourage efficiency and robustness. The newer works help develop more robust and context-aware relational reasoning towards open-world visual scenarios.

\begin{figure}
\centering
\begin{tikzpicture}[
  font=\fontfamily{ptm}\selectfont\small,
  obj/.style={circle, draw=black!55, line width=0.8pt, minimum size=1.1cm, align=center, inner sep=1pt},
  re/.style={-latex, line width=0.8pt, draw=black!60}
]
  \node[obj, fill=catTasks] (A) at (0,0)    {Obj~A};
  \node[obj, fill=catMeth]  (B) at (3,1.2)  {Obj~B};
  \node[obj, fill=catTypes] (C) at (3,-1.2) {Obj~C};
  \node[obj, fill=catEval]  (D) at (6,0)    {Obj~D};
  \draw[re] (A) -- node[above,sloped,font=\fontfamily{ptm}\selectfont\scriptsize]{rel1} (B);
  \draw[re] (A) -- node[below,sloped,font=\fontfamily{ptm}\selectfont\scriptsize]{rel2} (C);
  \draw[re] (B) -- node[above,sloped,font=\fontfamily{ptm}\selectfont\scriptsize]{rel3} (D);
  \draw[re] (C) -- node[below,sloped,font=\fontfamily{ptm}\selectfont\scriptsize]{rel4} (D);
\end{tikzpicture}
\caption{Relational reasoning over a scene graph: objects are nodes and labeled edges are pairwise relations. Multi-hop inference propagates along A$\rightarrow$B$\rightarrow$D and A$\rightarrow$C$\rightarrow$D.}
\label{fig:scene-graph-reasoning}
\end{figure}

\subsection{Causal Reasoning in Vision}

Whereas overall vision research centers on discerning correlations, causal reasoning tries to discern why things occur and how outcomes would be different under different circumstances \cite{chen2024learning,fang2024lowrank}. Central to this paradigm is the Structural Causal Model (SCM)~\cite{pearl2009causality}, which stores variable dependencies in directed acyclic graphs (DAGs) and enables both interventional and counterfactual inference. SCMs are not the only option. Causal structure in vision can equally be expressed with Bayesian networks and, more broadly, probabilistic graphical models~\cite{pearl1988probabilistic, koller2009pgm}, which share the DAG substrate but emphasize probabilistic inference over observed evidence. We adopt the SCM formulation here because of its explicit $do(\cdot)$-operator semantics for interventions and counterfactuals. As seen in Figure~\ref{fig:causal-graph-scm}, variables are nodes related by directed structural mechanisms. The red intervention arrow $do(X{=}x)$ sets a variable externally, the red cross marks the edge deleted under intervention, and the dashed query denotes a counterfactual outcome, which enables reasoning about how changing a cause ripples through the graph.

In computer vision, techniques such as CausalVQA~\cite{yang2022causalvqa} posit SCM principles on visual question answering, disentangling causal from observational data and inferring counterfactual intervention effect. Others include causal representation learning for feature disentangling, counterfactual data augmentation to prevent spurious correlations, and domain generalization approaches such as invariant risk minimization (IRM) to boost robustness during distribution shift. Although promising, they are challenged by dataset bias, unobserved confounders, and lack of causal annotations, and standard benchmarks such as CLEVR-Causal, VCR-Causal are synthetic/constrained and lack path interpretability of causation in an open world. More complications arise in dynamic tasks such as video reasoning, where lagged causal effects and multiple concurrent interventions complicate inference further.

A growing body of work seeks to overcome these shortcomings by integrating causal discovery with self-supervised representation learning, generating counterfactual observations, and injecting multimodal causal information from language inputs or sensor observations, all of which are steps toward mechanism-aware reasoning in dynamic, safety-critical scenarios.

\begin{figure}
\centering
\begin{tikzpicture}[
  font=\fontfamily{ptm}\selectfont\small,
  v/.style={circle, draw=black!60, line width=0.8pt, minimum size=1.0cm, align=center, inner sep=1pt},
  e/.style={-latex, line width=0.9pt, draw=black!65},
  ex/.style={-latex, line width=0.7pt, draw=black!45, dashed}
]
  \node[v, fill=catTasks] (X) at (0,0)   {$X$};
  \node[v, fill=catMeth]  (M) at (2.8,0) {$M$};
  \node[v, fill=catEval]  (Y) at (5.6,0) {$Y$};
  \node[font=\fontfamily{ptm}\selectfont\scriptsize] (Ux) at (0,1.5)   {$U_X$};
  \node[font=\fontfamily{ptm}\selectfont\scriptsize] (Uy) at (5.6,1.5) {$U_Y$};
  \draw[ex] (Ux)--(X); \draw[ex] (Uy)--(Y);
  \draw[e] (X)-- node[above,font=\fontfamily{ptm}\selectfont\scriptsize]{$f_M$} (M);
  \draw[e] (M)-- node[above,font=\fontfamily{ptm}\selectfont\scriptsize]{$f_Y$} (Y);
  \draw[e] (X) to[bend right=40] (Y);
  \node at (2.8,-1.15) {\textcolor{red}{\ding{55}}};
  \draw[-latex, line width=1pt, draw=red!75] (-2.1,0) -- node[above,font=\fontfamily{ptm}\selectfont\scriptsize,text=red!75]{$do(X{=}x)$} (X.west);
  \node[font=\fontfamily{ptm}\selectfont\scriptsize, draw=black!45, dashed, rounded corners, inner sep=2.5pt, fill=white, below=6mm of Y] (cf) {counterfactual: $Y_{do(X=x')}$?};
  \draw[ex] (Y)--(cf);
\end{tikzpicture}
\caption{Structural causal model (SCM) for visual reasoning. Solid edges are structural mechanisms ($X\!\to\!M\!\to\!Y$ and a direct $X\!\to\!Y$). The red arrow performs the intervention $do(X{=}x)$, the red \ding{55} marks the edge removed under intervention, and the dashed box poses a counterfactual query $Y_{do(X=x')}$. $U_X,U_Y$ denote exogenous noise.}
\label{fig:causal-graph-scm}
\end{figure}

\subsection{Modular and Sequential Reasoning Architectures}

Modular architectures attempt to divide big visual reasoning tasks into sub-problems that are better comprehended, each of which is addressed via a special-purpose module. Outputs of such modules are then sequenced or hierarchically assembled to construct an overall reasoning trail. In this formulation, modular architectures are amenable to compositional generalization and afford step-by-step interpretability \cite{li2023variational} that makes this especially attractive for tasks with explicit multi-step reasoning, and human-auditable states in between.

A characteristic example is MACNet~\cite{hudson2018compositional}, which uses a recurrent controller with an explicit memory to consecutively attend to relevant question components and image regions. For example, for a CLEVR-style question like “Is the cube left of the sphere red?”, the first reasoning step would localize the cube, the second would decide its spatial relation to the sphere, and the third would verify its color attribute, resulting in a transparent reasoning path. Each cognitive operation corresponds to a discrete reasoning step, facilitating thorough examination of the reasoning process. In addition to MACNet, related approaches are Neural Module Networks~\cite{hu2017learning}, which compose task-specific networks on-the-fly from learned modules, and memory-augmented transformers~\cite{rae2020compressive}, which preserve long-range dependencies and provide feedback-driven reasoning.

As tabulated in Table~\ref{tab:reasoning-models}, modular pipelines are state-of-the-art for multi-hop VQA in synthetic environments (e.g., CLEVR), yet memory-augmented transformers are preferred for temporally extended reasoning on video sequences. Modular networks do their best when explicit reasoning traces are necessary, while more adaptive forms like memory-augmented transformers more readily accommodate noisy, unstructured inputs. Although modular systems are explainable, their robustness and scalability are an issue. Modular systems typically require hand-curated supervision or program annotations and are thus less compatible with open-world data. Moreover, hard modular architectures become rigid in the presence of unexpected reasoning patterns during training time. Future research explores differentiable routing and soft module parameterization to make hard module boundaries softer, and uses large pretrained vision–language models (e.g., Flamingo, BLIP-2) as general-purpose perception backbones.

A notable recent shift is the migration from \emph{fixed} modular pipelines toward the \emph{agentic} paradigm, in which a large (multimodal) language model acts as a controller that dynamically plans, selects, and invokes external tools (detectors, segmenters, OCR, code execution, retrieval) at inference time rather than executing a pre-wired module graph~\cite{xi2025llmagents, shen2023hugginggpt}. In vision, VisProg~\cite{gupta2023visual} and ViperGPT~\cite{suris2023vipergpt} instantiate this idea by having the controller synthesize an executable program over visual primitives. The two perspectives are complementary: classical modular networks fix the module set and learn the routing, whereas agentic systems fix a general controller and let it compose tools on the fly, trading the trainability and verifiability of the former for the flexibility and open-vocabulary coverage of the latter, at the cost of weaker guarantees on the faithfulness and reproducibility of the resulting reasoning trace. A future direction is integrating modular reasoning with causal discovery or knowledge-grounded priors, enabling adaptive composition that better generalizes to open-world reasoning and zero-shot compositional tasks.

\begin{table}[H]
\footnotesize
\centering
\caption{Unified comparison of visual reasoning models by input modality, architecture, reasoning types, core tasks, and representative models.}
\label{tab:reasoning-models}
\setlength{\tabcolsep}{3pt}
\rowcolors{2}{rowblue}{white}% 从第2行起，浅蓝/白交替
\resizebox{\textwidth}{!}{%
\begin{tabular}{p{2.8cm} p{2.6cm} p{3.5cm} p{4.5cm} p{4.5cm}}
% \begin{tabular}{lllll}
\toprule
\rowcolor{headerblue}
\textbf{Input Modality} & \textbf{Architecture} & \textbf{Reasoning Type(s)} & \textbf{Core Task(s)} & \textbf{Representative Models} \\
\midrule
Image + Text & Neuro-symbolic pipeline & Symbolic, compositional & Visual QA, CLEVR-style logical reasoning & NS-CL~\cite{mao2019neuro}, CLEVR-Hans~\cite{stammer2021clevr} \\
Image & GNNs, Scene Graphs & Relational, spatial & Scene graph generation, object interaction understanding & Visual Genome GCN~\cite{yang2018graph}, MoTIFNet~\cite{zellers2018neural} \\
Image + Video & SCM-based causal models & Causal, counterfactual & Causal reasoning, VQA under interventions & CausalVQA~\cite{yang2022causalvqa}, SCM-Vision~\cite{bengio2019meta} \\
Video + Text & Spatio-temporal transformers & Temporal, causal & Action recognition, temporal question answering & TimeSformer~\cite{bertasius2021space}, TVQA~\cite{lei2018tvqa} \\
Image + Question & Modular networks (e.g., MACNet) & Sequential, compositional & Multi-hop VQA in synthetic scenes & MACNet~\cite{hudson2018compositional}, NS-VQA~\cite{yi2018neural} \\
Image & Logic-based reasoning & Symbolic, rule-based & Visual program execution, concept grounding & DeepProbLog~\cite{manhaeve2018deepproblog} \\
Image & Relation Networks & Pairwise relational reasoning & Comparative QA, Object Relationship Tasks & RN~\cite{santoro2017simple} \\
Image + Scene Graph & Graph Attention Networks & Hybrid (symbolic + relational) & Entity linking, semantic relation inference & GAT~\cite{velivckovic2017graph}, GSNN~\cite{marino2017more} \\
Video + Text & Memory-Augmented Transformers & Temporal, commonsense & Future event prediction, video captioning & Compressive Transformer~\cite{rae2020compressive}, Frozen~\cite{wu2022multi} \\
Image + KB & Knowledge-integrated models & Commonsense, intent inference & Social reasoning, intention prediction & VisualCOMET~\cite{park2020visualcomet}, VCR~\cite{zellers2019recognition} \\
\bottomrule
\end{tabular}
}
\end{table}

% \vspace{1em}

From Table~\ref{tab:reasoning-models}, it is evident that symbolic and modular architectures dominate in tasks requiring interpretability and compositional generalization, while relational and causal models are better suited for dynamic, context-rich environments. Temporal and commonsense reasoning remain less mature, with most advances coming from integrating large-scale vision–language models as priors.

\subsection{Vision–Language Models and Emergent Multimodal Reasoning}
\label{subsec:vlm-reasoning}

The methodologies above were largely conceived around task-specific architectures. Since 2022, a distinct paradigm has emerged in which reasoning is performed \emph{on top of} large pretrained vision–language models (VLMs) and multimodal large language models (MLLMs). Models such as Flamingo~\cite{alayrac2022flamingo}, BLIP-2~\cite{li2023blip2}, LLaVA~\cite{liu2023llava}, InstructBLIP~\cite{dai2023instructblip}, Qwen-VL~\cite{bai2023qwenvl}, and proprietary systems such as GPT-4V~\cite{openai2023gpt4v} and Gemini~\cite{geminiteam2023gemini} couple a visual encoder to a language model, exposing broad world knowledge and in-context learning to visual inputs. Rather than encoding a fixed inductive bias for one reasoning type, these models act as a general substrate on which relational, causal, temporal, and commonsense reasoning can be elicited through prompting, decoding strategies, or external tools. This reframes the methodological question from \emph{which architecture to design} to \emph{how to elicit and verify reasoning} from a pretrained model.

\textbf{Prompted and chain-of-thought reasoning.} Building on chain-of-thought (CoT) prompting in language models~\cite{wei2022chain}, multimodal CoT decomposes a visual query into intermediate textual rationales before producing an answer, improving multi-hop and compositional performance~\cite{zhang2023multimodalcot}. Recent work makes the rationale explicitly visual: Visual CoT collects grounded reasoning chains and benchmarks step-by-step inference~\cite{shao2024visualcot}, while LLaVA-CoT~\cite{xu2024llavacot} and Insight-V~\cite{dong2025insightv} train MLLMs to emit long, structured reasoning traces and demonstrate that \emph{test-time} reasoning, that is, allocating more inference-time computation to deliberate, multi-step chains, can substantially raise accuracy on reasoning-intensive benchmarks. Grounding-oriented prompts such as Set-of-Mark~\cite{yang2023setofmark} and grounded MLLMs such as Kosmos-2~\cite{peng2023kosmos2} further tie linguistic reasoning steps to specific image regions, partially restoring the spatial traceability that purely textual rationales lack.

\textbf{Tool use and visual programming.} A complementary line treats the (M)LLM as a controller that composes external visual modules, reconnecting the foundation-model paradigm with the modular and neuro-symbolic traditions discussed above. VisProg~\cite{gupta2023visual} and ViperGPT~\cite{suris2023vipergpt} prompt an LLM to synthesize an executable program that calls detectors, segmenters, and VQA primitives, yielding an explicit, inspectable reasoning trace without task-specific training. Chameleon~\cite{lu2023chameleon} and MM-ReAct~\cite{yang2023mmreact} generalize this to plug-and-play tool orchestration and interleaved reasoning-and-acting. Because the intermediate program or tool trace is symbolic, these systems inherit much of the auditability of classical neuro-symbolic pipelines while leveraging the open-vocabulary perception of foundation models, at the cost of dependence on prompt quality and the reliability of the underlying tools.

\textbf{Positioning within the taxonomy and open limitations.} VLM/MLLM-based methods cut across all five reasoning types: relational and symbolic reasoning surface through generated programs and scene descriptions, temporal reasoning through video-instruction tuning, causal reasoning through counterfactual prompting, and commonsense reasoning through the language model's pretrained priors. However, this breadth comes with distinctive failure modes that classical methods did not exhibit at the same scale. First, \emph{faithfulness}: a fluent rationale or program need not reflect the computation that produced the answer, so a correct final answer can mask incorrect or hallucinated intermediate steps. Second, \emph{language-prior shortcuts}: strong text priors let models answer many ``visual'' questions without genuinely attending to the image, inflating benchmark scores. Third, \emph{evaluation and reproducibility}: closed weights, prompt sensitivity, benchmark contamination, and capability-specific stress tests such as selective forgetting~\cite{yu2025forgetme} make controlled comparison difficult. Fourth, \emph{cost and latency}: long visual reasoning chains and multi-tool pipelines are expensive, complicating deployment in the real-time, resource-constrained settings emphasized throughout this survey. These observations motivate the faithfulness-, structure-, and causality-aware evaluation protocols discussed in Section~\ref{sec:evaluation}, which are precisely the dimensions on which accuracy-only benchmarks fail to separate genuine reasoning from sophisticated pattern matching.

\subsection{Comparative Analysis and Trends}

As shown in Tables~\ref{tab:xai_cv_reasoning} and \ref{tab:reasoning-models}, symbolic methods are best at explainability but need highly structured input, neural models provide flexibility but at the cost of explainability, hybrid frameworks try to have the best of both but suffer from integration overhead, and modular frameworks give compositional reasoning but are poor at scalability in unstructured environments. Recent work shows an overt tendency to integrate multiple paradigms, for example, mixing causal discovery into graph reasoning, injecting differentiable logic into transformers, or using large vision–language models to ground commonsense.

Here, we have pinpointed three promising future research trajectories, namely (i) cross-paradigm fusion pipelines with symbolic auditability and neural adaptability, (ii) efficiency- and scaling-aware architectures for high-resolution, temporally extended inputs, and (iii) generalization-oriented testing on diverse, realistic test suites. This unifying methodology is needed for next-generation visual reasoning systems runnable in open-world, safety-critical applications.

To make these trends concrete, Table~\ref{tab:quantitative} compiles representative reported results across five widely used reasoning benchmarks. Three observations stand out. First, on the synthetic, fully compositional CLEVR benchmark, the neuro-symbolic NS-VQA attains near-perfect accuracy (99.8\%), surpassing strong neural and modular models, which is evidence that explicit symbolic structure pays off when the visual domain is clean and the program space is well defined. Second, on the more naturalistic GQA and VQA~v2, large pretrained transformers and VLMs/MLLMs (LXMERT, BLIP, LLaVA-1.5) dominate, reflecting the value of broad pretraining for open-world perception and language priors. However, the gap between BLIP-2's zero-shot VQA~v2 accuracy (65.0\%) and the fine-tuned models (up to 80.0\%) shows that much of this apparent competence still depends on in-domain supervision—and that headline VQA accuracy partly reflects strong language priors rather than visual reasoning per se. Third, and most tellingly, the CLEVRER rows expose how thin current ``reasoning'' often is: even the best neuro-symbolic model drops from 74.1\% on \emph{predictive} questions to 42.2\% on \emph{counterfactual} questions, and purely neural baselines collapse to near chance (25.1\%). The gap between high aggregate accuracy and weak counterfactual performance is exactly the failure mode that accuracy-only leaderboards hide, motivating the structural and causal evaluation dimensions developed in Section~\ref{sec:evaluation}.

\begin{table}[H]
\centering
\caption{Representative reported results on five visual-reasoning benchmarks, spanning neural, modular, neuro-symbolic, and VLM/MLLM paradigms. Numbers are as reported by the cited sources and are \emph{not} strictly comparable across rows due to differing splits, supervision, and pretraining data. They are intended to illustrate trends, not to rank models. \textsuperscript{a}30-model ensemble, test-standard. \textsuperscript{b}test-dev; VQA~v2 training images were seen during instruction tuning. \textsuperscript{c}per-question accuracy on predictive\,/\,counterfactual questions. \textsuperscript{d}validation Q$\rightarrow$A\,/\,Q$\rightarrow$AR. \textsuperscript{e}LLaVA-1.5 observes the GQA training split during instruction tuning (in-domain, not zero-shot). \textsuperscript{f}zero-shot (no VQA~v2 fine-tuning), test-dev; shown to illustrate the zero-shot gap.}
\label{tab:quantitative}
\footnotesize
\renewcommand{\arraystretch}{1.15}
\begin{tabular}{@{}>{\raggedright\arraybackslash}p{2.0cm} >{\raggedright\arraybackslash}p{2.7cm} >{\raggedright\arraybackslash}p{2.7cm} c c@{}}
\toprule
\rowcolor{headerblue}
\textbf{Benchmark} & \textbf{Model} & \textbf{Paradigm} & \textbf{Acc.\ (\%)} & \textbf{Year} \\
\midrule
\rowcolor{rowblue} CLEVR        & CNN+LSTM+SA~\cite{johnson2017inferring} & Neural (attention)      & 69.8 & 2017 \\
\rowcolor{rowblue} (overall)    & IEP (PG+EE)~\cite{johnson2017inferring} & Modular / program       & 96.9 & 2017 \\
\rowcolor{rowblue}              & MAC~\cite{hudson2018compositional}      & Modular / attention     & 98.9 & 2018 \\
\rowcolor{rowblue}              & NS-VQA~\cite{yi2018neural}              & Neuro-symbolic          & \textbf{99.8} & 2018 \\
\addlinespace
GQA          & MAC~\cite{hudson2019learning}           & Modular / attention     & 54.1 & 2019 \\
(test)       & LXMERT~\cite{tan2019lxmert}             & Transformer VLM         & 60.3 & 2019 \\
             & NSM~\cite{hudson2019learning}           & Neuro-symbolic / graph  & \textbf{63.2} & 2019 \\
             & LLaVA-1.5-13B\textsuperscript{e}~\cite{liu2023improvedllava} & MLLM      & 63.3 & 2023 \\
\addlinespace
\rowcolor{rowblue} VQA~v2       & BUTD\textsuperscript{a}~\cite{anderson2018bottom} & Neural (attention) & 70.3 & 2018 \\
\rowcolor{rowblue} (overall)    & LXMERT~\cite{tan2019lxmert}             & Transformer VLM         & 72.5 & 2019 \\
\rowcolor{rowblue}              & BLIP-2\textsuperscript{f}~\cite{li2023blip2} & VLM (zero-shot)     & 65.0 & 2023 \\
\rowcolor{rowblue}              & BLIP~\cite{li2022blip}                  & VLM (pretrained)        & 77.6 & 2022 \\
\rowcolor{rowblue}              & LLaVA-1.5-13B\textsuperscript{b}~\cite{liu2023improvedllava} & MLLM       & \textbf{80.0} & 2023 \\
\addlinespace
CLEVRER      & MAC (V+)\textsuperscript{c}~\cite{yi2020clevrer} & Neural (attention) & 63.5 / 25.1 & 2020 \\
(pred./c.f.) & NS-DR\textsuperscript{c}~\cite{yi2020clevrer}    & Neuro-symbolic     & \textbf{74.1} / \textbf{42.2} & 2020 \\
\addlinespace
\rowcolor{rowblue} VCR          & R2C\textsuperscript{d}~\cite{zellers2019recognition} & Neural (grounded) & 63.8 / 43.1 & 2019 \\
\rowcolor{rowblue} (val)        & UNITER-large\textsuperscript{d}~\cite{chen2020uniter} & Transformer VLM  & \textbf{77.3} / \textbf{62.8} & 2020 \\
\bottomrule
\end{tabular}
\end{table}

\section{Evaluation}
\label{sec:evaluation}

Testing visual reasoning models is much more than testing output accuracy. It tests \textbf{what} is predicted, \textbf{how}, and \textbf{why} it holds, across changing conditions. Unlike typical classification, reasoning problems include multi-step reasoning, relational structure, and even causal processes, all aspects that scalar summary scores cannot capture. Current evaluation methods frequently overfit to short, narrow benchmarks, with unresolved issues related to robustness, generalizability, and faithfulness to desired reasoning processes.

To address these gaps, we adopt a three-pronged framework:

\begin{enumerate}
    \item \textbf{Functional correctness} is the model generating the correct result on compositional, multi-hop, and logic-rich tasks?
    \item \textbf{Structural consistency} are the intermediate reasoning paths coherent and interpretable?
    \item \textbf{Causal validity} does the model capture cause–effect relations and react appropriately to interventions?
\end{enumerate}

\subsection{Functional Metrics}

Functional evaluation examines whether a model produces correct outputs in tasks that demand compositional, multi-hop, or logical inference. While simple cases can be assessed using standard accuracy, more challenging settings require metrics tailored to reasoning complexity.

Benchmarks such as CLEVR~\cite{johnson2017clevr} and GQA~\cite{hudson2019gqa} explicitly embed multi-step reasoning requirements, enabling fine-grained error analysis. Diagnostic datasets like NLVR2~\cite{suhr2019corpus} and ACRE~\cite{andreas2022acre} introduce controlled visual–linguistic perturbations to test robustness under distribution shifts.

A representative measure is the compositional generalization accuracy:
\begin{equation}
\text{Accuracy}_{\text{comp}} = \frac{\text{Correct Answers in Novel Compositions}}{\text{Total Novel Composition Samples}}
\end{equation}
which quantifies a model’s ability to generalize to unseen combinations of familiar concepts, a hallmark of genuine reasoning. Beyond overall scores, functional evaluation often decomposes performance into subtasks (e.g., attribute comparison, set operations, object counting), yielding interpretable insights at the module or capability level.

Nevertheless, prevailing functional metrics tend to function in artificial or highly controlled settings, potentially overstating robustness in the everyday world. Most tasks don't embody reasoning about noisy, incomplete, or adversarial visual observations, and compositional generalization tends to only be evaluated under limited concept vocabularies. Consequently, excellent performance on benchmark collections might not generate robust reasoning under uncontrolled, everyday situations, which points to the need to develop evaluation protocols involving more task variety, ecologies of natural data, and inter-domain tests of generalization.

It is worth noting that these functional notions are not new: the distinction between global and partial \emph{correctness}, and the dual notions of \emph{coverage} and \emph{completeness}, have been formalized for decades in knowledge representation and reasoning and in ontology/knowledge-base evaluation~\cite{brank2005ontologyeval, cropper2022ilp30}. Reading visual-reasoning functional metrics through this lens clarifies what they do and do not certify. For instance, per-step correctness is a soundness check on a single derivation, whereas compositional generalization probes coverage over a concept space. It also exposes a trade-off, in that metrics that reward completeness over a broad vocabulary tend to tolerate locally incorrect steps, while strict step-correctness metrics say little about coverage. In addition to compositional generalization, functional assessment can also include quantities like the reasoning-step success rate, which measures the ratio of correct intermediate reasoning steps in multi-hop chains, and factual consistency scores, which estimate alignment between predicted responses and verifiable world knowledge in knowledge-grounded reasoning. Strength-based functional measures further investigate a model’s robustness against controlled disturbances, like object occlusion, distractor addition, or modality dropout, demonstrating resistance to visual–linguistic noise. The selection of functional measures is task-dependent per se: single-hop, closed-domain reasoning can be properly evaluated with typical accuracy, but multi-hop synthetic tasks are aided by per-step correctness and compositional generalization accuracy, and open-domain/cross-modal tasks need factual consistency and robustness measures. Individually, these refinements underscore that functional correctness is not an all-or-nothing property but a continuum of task-dependent abilities, each illuminating different vulnerabilities/strengths of a visual reasoning system.

\subsection{Structural Metrics}

Whereas functional metrics only measure terminal outputs, structural evaluation looks at the reasoning process, asking if internal workings of the model consist of well-defined and understandable sequential steps. This requires inspection of intermediate representations, like attention maps, scene graphs, or module activities, in order to check if their content is consistent with the desired reasoning path. For instance, for graph-based models, faithfulness of predicted scene graphs to ground truth annotations can be evaluated via similarity measures like:
\begin{equation}
\text{GraphSim}(G_1, G_2) = \frac{|E_1 \cap E_2|}{|E_1 \cup E_2|}
\end{equation}
where \(E_1\) and \(E_2\) denote the edge sets of the predicted and reference graphs. A higher score indicates greater relational reasoning fidelity. Here too the underlying machinery predates visual reasoning: graph similarity, graph edit distance, and global/local consistency have been studied extensively in graph matching for pattern recognition and in knowledge-graph and semantic-web evaluation~\cite{conte2004graphmatching, gao2010ged, wang2017kgembedding}. The simple Jaccard-style overlap above is a coarse special case. Borrowing richer, structure-aware similarity and consistency measures from that literature would make structural evaluation more discriminative and less dependent on exact node and edge alignment. Similarly, architectures like MACNet~\cite{hudson2018compositional} produce explicit multi-step reasoning traces, allowing evaluations to check whether module activations correspond to the expected logical sequence. Beyond these, structural assessment has expanded to include attention faithfulness, module relevance, and emerging approaches such as concept graph alignment or disentangled latent structures~\cite{boger2023conceptgraph}, which enhance interpretability by making the reasoning process more transparent.

Recent developments have broadened the scope of structural evaluation. Multimodal alignment measures assess whether intermediate representations in vision--language models remain consistent across modalities. Logic-chain consistency tests apply formal reasoning or symbolic execution to verify the validity of intermediate steps, while process-based probing introduces controlled interventions at specific reasoning stages to diagnose model behavior. Other robustness-oriented evaluations test the stability of intermediate structures under noise injection, pruning, or adversarial perturbations, thereby examining whether the reasoning path is resilient rather than brittle.

Even with all these developments, however, structural metrics have inherent limitations. Many make simplistic assumptions that an intermediate representation selected for attention mapping (e.g., an attention map, scene graph) faithfully indicates the model’s reasoning, an assumption that won’t necessarily hold true in practice~\cite{samek2019explainable}. The attention weights can be manipulated without changing predictions, the scene graph labels are usually incomplete or subjective, and the majority of structural assessments are on tiny, hand-curated sets with limited ecological validity. A high structural score may therefore indicate conformity to a proxy representation instead of real reasoning capability. To counteract this, we need multi-view structural verification that cross-checks attention maps, scene graphs, and execution traces, together with scaling assessments to large, real-world sets and the addition of causal analysis to make sure that intermediary traces reflect underlying mechanisms instead of spurious correlations.

\subsection{Causal Metrics}

Causal evaluation extends reasoning analysis from correlation, attempting to verify if a model is capable of discerning, modeling, and responding to cause–effect relationships within visual information. Causal reasoning is distinct from strictly correlational approaches in that it directly assesses a model’s responses to interventions and its ability to return valid counterfactuals. Causal evaluation, in additional advanced scenarios, also assesses if related cause–effect knowledge can be transferred from one task or domain to another, which would reveal a model’s ability to generalize causal patterns, not overfit to one environment.

A widely used measure is the Average Causal Effect (ACE):
\begin{equation}
\text{ACE} = \mathbb{E}[Y \mid do(X = x_1)] - \mathbb{E}[Y \mid do(X = x_0)]
\end{equation}
which quantifies the expected change in the outcome \( Y \) when intervening to set variable \( X \) from \( x_0 \) to \( x_1 \).

Another key measure is the Natural Direct Effect (NDE):
\begin{equation}
\text{NDE} = \mathbb{E}[Y_{X=1, M=M_{X=0}}] - \mathbb{E}[Y_{X=0}]
\end{equation}
where \( M \) represents a mediator variable, enabling isolation of direct causal influence by controlling for indirect effects.

\begin{figure}
    \centering
    \includegraphics[width=0.7\linewidth]{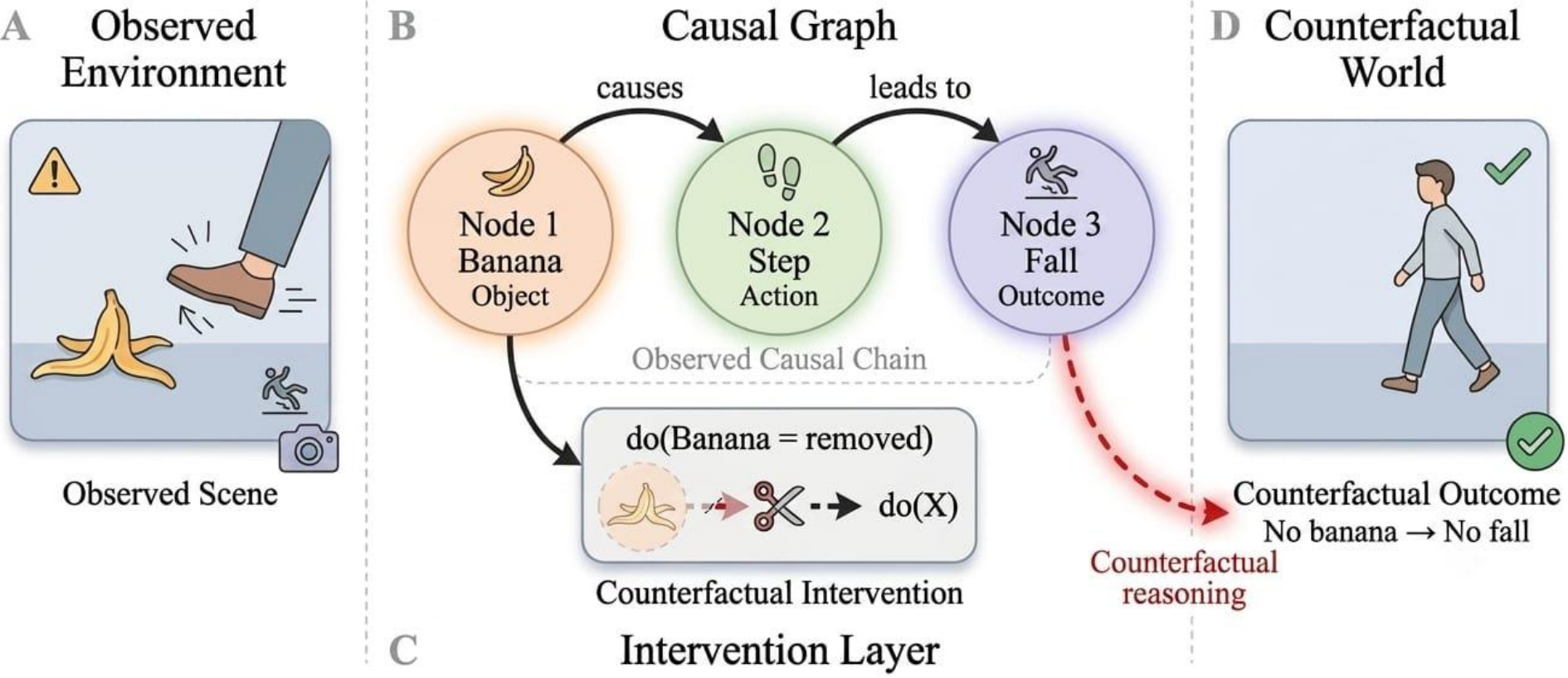}
    \caption{Causal DAG for a ``slip-and-fall'' scene: \textit{Banana} $\rightarrow$ \textit{Step} $\rightarrow$ \textit{Fall}. The dashed red edge denotes a counterfactual intervention, where removing the banana should break the chain and prevent the fall.}
    \label{fig:causal-dag}
\end{figure}

These metrics, grounded in structural causal models (SCM)~\cite{pearl2009causality}, have been applied in visual reasoning benchmarks such as CausalVQA~\cite{yang2022causalvqa} and the Backgrounds Challenge~\cite{lopez2023imagenetcausal}. Figure~\ref{fig:causal-dag} illustrates a causal DAG example (\textit{Banana} $\rightarrow$ \textit{Step} $\rightarrow$ \textit{Fall}) that supports interventional and counterfactual queries.

An equally important dimension is counterfactual consistency: whether a model revises its predictions appropriately when a key causal factor is altered. For instance, in a “slip-and-fall” video, removing the banana should alter the predicted outcome. Synthetic datasets such as CLEVRER~\cite{yi2020clevrer} provide controlled manipulations for systematically testing this property. Some newer benchmarks also introduce dynamic interventions in simulated environments (e.g., physics engines) to assess whether models adapt their reasoning when causal dependencies change over time.

However, despite their diagnostic value, such datasets often lack the complexity of real-world causality, leaving robustness in unconstrained environments an open challenge. Most causal evaluation procedures are critically built on top of simplified scenarios with precisely defined causal graphs, which are only exceptionally representative of the ambiguity, confounding, and feedback of true vision tasks in the wild. Natural-scenes ground-truth causal structures are typically unobservable, and so these approaches are bound to rely on synthetically-generated benchmarks potentially overstating a model’s generalization to uncontrolled environments. Additionally, intervention in high-dimensional vision data is non-trivial, since pixel-level changes may spuriously introduce relevant cues, so that an observed “causal effect” measured may represent spurious sensitivity to low-level changes rather than true causal reasoning. These constraints underlie the disconnect of today’s causal metrics and the challenge of robust, in-the-wild causal inference in vision.

% \begin{figure}
% \centering
% \begin{tikzpicture}[
%   font=\fontfamily{ptm}\selectfont\small,
%   n/.style={rounded corners=3pt, draw=black!60, line width=0.8pt, align=center, inner sep=5pt, text width=2.2cm, minimum height=1.1cm},
%   e/.style={-latex, line width=0.9pt, draw=black!65}
% ]
%   \node[n, fill=catMeth]  (b) at (0,0)   {\textbf{Banana}\\[1pt]{\scriptsize object}};
%   \node[n, fill=catTypes] (s) at (3.7,0) {\textbf{Step}\\[1pt]{\scriptsize action}};
%   \node[n, fill=catEval]  (f) at (7.4,0) {\textbf{Fall}\\[1pt]{\scriptsize outcome}};
%   \draw[e] (b) -- node[above,font=\fontfamily{ptm}\selectfont\scriptsize]{causes} (s);
%   \draw[e] (s) -- node[above,font=\fontfamily{ptm}\selectfont\scriptsize]{leads to} (f);
%   \draw[-latex, dashed, draw=red!75, line width=0.8pt] (f) to[bend left=32]
%         node[below,font=\fontfamily{ptm}\selectfont\scriptsize,text=red!75]{counterfactual: no banana $\Rightarrow$ no fall?} (b);
% \end{tikzpicture}
% \caption{Causal DAG for a ``slip-and-fall'' scene: \textit{Banana} $\rightarrow$ \textit{Step} $\rightarrow$ \textit{Fall}. The dashed red edge denotes a counterfactual intervention, where removing the banana should break the chain and prevent the fall.}
% \label{fig:causal-dag}
% \end{figure}

Aside from regular accuracy, assessment protocols also validate counterfactual consistency, i.e., does a model revise its predictions correctly once a principal causal variable has been modified? As summarized in Table~\ref{tab:evaluation-metrics}, no single benchmark comprehensively captures all reasoning types or supervision settings. Functional metrics measure outcome correctness, structural evaluations probe the internal reasoning process, and causal assessments validate a model’s capacity for true inferential understanding. But existing metrics tend to concentrate on slim types of reasoning, synthetic situations, or sparse modalities such that cross-paradigm and real-world generalization remains an open issue. A comprehensive testing of visual reasoning thus requires a multi-faceted scheme that combines functional, structural, and causal views under realistic scenarios.

\begin{table}[H]
\centering
\footnotesize
\caption{Evaluation dimensions and metrics for visual reasoning models.}
\label{tab:evaluation-metrics}
\renewcommand{\arraystretch}{1.2}
\rowcolors{2}{rowblue}{white}% 从第2行起，浅蓝/白交替
\begin{tabularx}{\linewidth}{p{3cm}X}
\toprule
\rowcolor{headerblue}
\textbf{Evaluation Type} & \textbf{Representative Metrics and Criteria} \\
\midrule
Functional Evaluation &
Task Accuracy, Multi-step Reasoning Accuracy, Compositional Generalization Accuracy \\
Structural Evaluation &
GraphSim (Graph Similarity), Attention Map Alignment, Module Trace Faithfulness \\
Causal Evaluation &
Average Causal Effect (ACE), Natural Direct Effect (NDE), Counterfactual Consistency Score \\
\bottomrule
\end{tabularx}
\end{table}

\subsection{Toward Unified Reasoning Evaluation}

Cross-dimensional integration is essential: high functional scores and low structural fidelity can be a sign of shortcut exploitation, and robust causal measures without functional accuracy can be an indication of theoretical but non-viable reasoning. Future assessment needs to transition to multi-view reasoning diagnostics, wherein functional, structural, and causal measures are exercised collectively, uncovering failure modes and trade-offs.
Challenging research directions for the future are threefold. The first is constructing ecologically valid, multimodal reasoning standards with causally justified structures. The second is constructing adaptive assessment pipelines whose tasks adapt adversarially to close shortcut loopholes. The third is combining process-based metrics (structural and causal faithfulness) with training objectives, building feedback loops that improve both accuracy and interpretability.

\section{Challenges}
\label{sec:challenges}

Despite rapid progress, current visual reasoning models still face limitations that hinder real-world deployment. Open problems remain in scalability, interpretability, generalization, and supervision. This section discusses four such challenges and the research directions that could make systems more robust and better aligned with human reasoning.

\subsection{Scalability and Generalization}

Most visual reasoning models exhibit performance degradation when faced with complex, high-resolution, or temporally extended inputs. Their computational demands make them unsuitable for real-time or resource-limited applications such as robotics or mobile devices~\cite{dosovitskiy2020image, bertasius2021space}. Moreover, models trained on synthetic benchmarks like CLEVR~\cite{johnson2017clevr} often fail to generalize to real-world scenarios, where visual and linguistic variations are more diverse and unpredictable.

Addressing these concerns requires developing architectures that are both scalable and modular. Recent efforts include hierarchical reasoning pipelines, sparse transformer variants like Longformer~\cite{beltagy2020longformer}, retraining-free token pruning for vision transformers~\cite{shi2026efficient}, and continuous learning frameworks that adapt over time. Equally important is improving domain generalization. Unsupervised domain adaptation methods~\cite{yu2024capan,wang2024multi} offer one route to cross-domain transfer, though they seldom target compositional reasoning directly, and training models that can reason across heterogeneous datasets without retraining for each new task remains largely underexplored but essential for robust deployment in open-world environments.

Despite these advances, most “scalable” or “generalizable” architectures are validated only on controlled benchmarks and do not show consistent gains on unconstrained, real-world streams. The synthetic-to-real gap is concrete. The neuro-symbolic recipe that reaches \textbf{99.8\%} on the clean, fully compositional CLEVR benchmark (Table~\ref{tab:quantitative}) has no comparable analogue on naturalistic data, and even strong pretrained VLMs/MLLMs depend on in-domain supervision: on VQA~v2 the best instruction-tuned model (LLaVA-1.5) reaches \textbf{80.0\%}, whereas BLIP-2 evaluated zero-shot reaches only \textbf{65.0\%} (Table~\ref{tab:quantitative}). This suggests that much of the headline performance reflects language priors and in-domain tuning rather than transferable visual reasoning. Sparse attention and hierarchical designs often trade reasoning fidelity for efficiency, which degrades multi-step inference. Domain generalization methods often rely on surface distribution alignment and ignore deeper shifts in causal structure or task semantics. Without addressing these mismatches, scalability solutions risk producing models that look robust in the lab but break down in open-world deployment.

\subsection{Hybrid Reasoning Paradigms}

Neuro-symbolic reasoning, which integrates neural perception with symbolic logic, holds great promise for interpretable and generalizable systems~\cite{mao2019neuro, yi2018neural,dampfhoffer2024backpropagation,song2024model,zhang2024critical}. These hybrid models attempt to combine the adaptability of deep learning with the structured expressivity of symbolic representations. However, this integration is fraught with practical challenges: inconsistencies in data representation, difficulties in gradient propagation, and issues of computational tractability hinder widespread adoption.

Progress in this direction requires innovation on multiple fronts. One priority is designing differentiable symbolic operators that align with semantic structures extracted from neural encoders. Another is facilitating semantic alignment between neural embeddings and logical abstractions. Robustness to noise and better handling of domain shifts will also be essential. Future systems may benefit from incorporating external knowledge bases, enabling zero-shot or compositional reasoning capabilities.

Although neuro-symbolic methods are promised as “best of both worlds” hybrids, most state-of-the-art systems show only narrow-scope success, usually on carefully curated tests. The brittleness is measurable. On CLEVRER, the strongest neuro-symbolic video model (NS-DR) scores \textbf{74.1\%} on predictive questions but only \textbf{42.2\%} on counterfactual ones (Table~\ref{tab:quantitative}), a drop of about 32 points once genuine intervention, rather than pattern completion, is required. In practice, symbolic components are fragile under noisy or uncertain visual input, and neural components come to dominate, so the symbolic layer becomes a post-hoc filter rather than a reasoning engine. The integration cost, in engineering complexity and computational overhead, can also outweigh the interpretability gains in dynamic, real-time settings. Without broad testing in unstructured, cross-domain conditions, neuro-symbolic reasoning may stay a niche curiosity rather than a deployable approach.

\subsection{Benchmark Limitations}

Many widely used datasets evaluate reasoning in narrowly scoped or artificially constructed environments. CLEVR, while compositional, lacks real-world complexity. GQA and VCR provide more naturalistic images, but cover a limited range of reasoning skills. Multilingual MLLM benchmarks such as MalayMMLU~\cite{poh2024malaymmlu} broaden language coverage yet still emphasize text-centric multitask evaluation rather than visual structural reasoning. As shown in Table~\ref{tab:heatmap-table}, no single benchmark adequately covers all types of reasoning or supervision scenarios.

To make Table~\ref{tab:heatmap-table} reproducible rather than impressionistic, we assign each benchmark a support score in $[0,1]$ for each reasoning type according to a fixed rubric that reflects how strongly the benchmark's \emph{task design and annotations require} that reasoning type: \textbf{1.0} the reasoning type is the benchmark's central design target (e.g., compositional symbolic chains in CLEVR, causal queries in CausalVQA); \textbf{0.6--0.9} substantially and explicitly required by a large fraction of items; \textbf{0.3--0.5} present as a secondary or implicit requirement; \textbf{0.1--0.2} only incidentally exercised; and \textbf{0.0} absent by construction. Scores were assigned independently by two of the authors from each benchmark's documentation and sample items, with disagreements (initial agreement on roughly four of five cells) resolved by discussion; the supervision column records the density of reasoning-relevant annotation (strong/moderate/weak). The scores are therefore a deliberately coarse, criterion-based map of coverage, not a measured model-performance quantity.

To advance the field, new datasets should combine realism with compositionality, include various types of reasoning, and offer multimodal inputs with varying levels of supervision. Rich intermediate annotations, such as causal graphs or reasoning traces, would also enable more informative structural and causal evaluations~\cite{yang2022causalvqa, park2020visualcomet}.

Dependence on benchmark-driven development today may generate instead models optimized for “benchmark gaming” rather than reasoning itself. Most data unwittingly bleed spurious correlations or short-cut cues, resulting in successful high scores rather than true multi-step inference. Furthermore, failure to regularize reasoning taxonomy and metric definitions themselves means cross-benchmark comparison cannot be trusted and thus cumulative progress frustrated. Until adversarially constructed benchmarks are embraced by the community as ecologically valid and ever-changing, visual reasoning research may be relegated to an evaluation bubble with impoverished translatability to open-world tasks.

% \vspace{0.5em}

% Color definitions
\definecolor{low}{RGB}{236,243,250}    % 极浅蓝
\definecolor{medium}{RGB}{198,219,239} % 浅蓝
\definecolor{high}{RGB}{107,174,214}   % 中蓝

% Heatcell macro
\newcommand{\heatcell}[1]{%
  \ifdim #1pt < 0.3pt\cellcolor{low} #1%
  \else\ifdim #1pt < 0.7pt\cellcolor{medium} #1%
  \else\cellcolor{high} #1%
  \fi\fi
}

\begin{table}[H]
\centering
\footnotesize
\caption{Reasoning-type support scores for major visual benchmarks, assigned with the $[0,1]$ rubric defined in the text (1.0 = central design target, 0.0 = absent). Color intensity indicates strength: red (low, $<0.3$), yellow (medium, $0.3$--$0.7$), and green (high, $\geq0.7$). Scores are a criterion-based coverage map, not measured model performance.}
\label{tab:heatmap-table}
% \resizebox{\textwidth}{!}{%
\begin{tabular}{l c c c c c c}
\toprule
\textbf{Dataset} & \textbf{Relational} & \textbf{Symbolic} & \textbf{Temporal} & \textbf{Causal} & \textbf{Commonsense} & \textbf{Supervision} \\
\midrule
CLEVR~\cite{johnson2017clevr}                         & \heatcell{1.0} & \heatcell{1.0} & \heatcell{0.0} & \heatcell{0.0} & \heatcell{0.0} & \cellcolor{high} Strong \\
CLEVR-X~\cite{salewski2022clevrx,unsal2025easyarc}    & \heatcell{1.0} & \heatcell{1.0} & \heatcell{0.0} & \heatcell{0.0} & \heatcell{0.2} & \cellcolor{medium} Moderate \\
GQA~\cite{hudson2019gqa,liu2023verify}                & \heatcell{0.9} & \heatcell{0.8} & \heatcell{0.2} & \heatcell{0.1} & \heatcell{0.3} & \cellcolor{high} Strong \\
VCR~\cite{zellers2019recognition,shangguan2024tomato} & \heatcell{0.3} & \heatcell{0.2} & \heatcell{0.6} & \heatcell{0.2} & \heatcell{1.0} & \cellcolor{medium} Moderate \\
VisualCOMET~\cite{park2020visualcomet,liu2023vsr}     & \heatcell{0.2} & \heatcell{0.1} & \heatcell{0.7} & \heatcell{0.2} & \heatcell{1.0} & \cellcolor{medium} Weak \\
CausalVQA~\cite{yang2022causalvqa,corbiere2025drivingvqa} & \heatcell{0.6} & \heatcell{0.3} & \heatcell{0.5} & \heatcell{1.0} & \heatcell{0.3} & \cellcolor{medium} Moderate \\
TVQA~\cite{lei2018tvqa,chen2025temporalvisualdyn}     & \heatcell{0.7} & \heatcell{0.2} & \heatcell{0.9} & \heatcell{0.3} & \heatcell{0.8} & \cellcolor{medium} Moderate \\
Salient ImageNet~\cite{singla2022salientimagenet,pandya2024ntsebench} & \heatcell{0.1} & \heatcell{0.0} & \heatcell{0.0} & \heatcell{1.0} & \heatcell{0.0} & \cellcolor{low} Weak \\
SSv2~\cite{goyal2017something,wang2025timecausality}  & \heatcell{0.8} & \heatcell{0.0} & \heatcell{1.0} & \heatcell{0.3} & \heatcell{0.2} & \cellcolor{low} Weak \\
VQA-v2~\cite{antol2015vqa,chen2024rextime,piryani2025temporalir} & \heatcell{0.3} & \heatcell{0.2} & \heatcell{0.1} & \heatcell{0.1} & \heatcell{0.3} & \cellcolor{high} Strong \\
\bottomrule
\end{tabular}
% }
\end{table}

\subsection{Weak Supervision}

A major bottleneck in visual reasoning is the reliance on strongly annotated data, such as scene graphs, object masks, and program templates. However, such annotations are costly and infeasible in domains such as remote sensing or healthcare, where labeling requires expert knowledge and time.

Recent approaches have begun to leverage weakly supervised and self-supervised signals. Contrastive learning~\cite{radford2021learning}, masked image modeling, few-shot prompting, and layered self-supervised knowledge distillation~\cite{dahri2026layered} are being adapted for reasoning tasks. Vision language models like BLIP and Flamingo~\cite{li2022blip} have shown that pre-trained cross-modal models can perform structured reasoning with minimal labeled data.
In the future, a key goal is to develop reasoning agents that learn from indirect supervision, such as rewards, textual feedback, or structural priors, and generalize across domains. Promising directions include symbolic bootstrapping from raw data, self-supervised logic induction, and cross-domain transfer of reasoning templates.

While weak supervision is often positioned as a solution to annotation bottlenecks, many existing “weakly supervised” setups still rely on hidden strong supervision, for example, large pre-training corpora with implicit human curation or datasets that leak label semantics into input features. This creates a misleading sense of generalization, as models may overfit to latent biases in the pre-training data rather than acquiring genuine reasoning skills. Moreover, evaluation under weak supervision is often inconsistent, lacking rigorous stress tests to verify robustness outside curated domains. Without standardized protocols that separate true weak supervision from concealed full supervision, the field risks overstating progress and underestimating the brittleness of current methods.

\section{Conclusion}

This survey examined how visual reasoning is moving computer vision from pattern recognition toward more cognitively inspired inference. It covered five core reasoning types, namely relational, symbolic, temporal, causal, and commonsense, and analyzed their representative models, datasets, and evaluation protocols within one unified framework. We placed the recent paradigm of reasoning with vision–language and multimodal large language models, including visual chain-of-thought, visual programming, and tool-augmented and test-time reasoning, alongside classical graph-based, neuro-symbolic, and modular methods, and showed how it re-expresses every reasoning type while adding new failure modes such as unfaithful rationales and language-prior shortcuts. Progress depends on accuracy, but equally on the faithfulness, interpretability, and causal validity of the reasoning process. Despite advances in graph neural networks, neuro-symbolic frameworks, spatio-temporal transformers, and foundation models, current systems still struggle with scalability, generalization to real-world data, integration of neural and symbolic reasoning, benchmark coverage, and reliance on strong supervision. Much of today’s progress remains benchmark-bound and skewed toward narrow, curated domains, with limited robustness in unconstrained settings. Addressing these bottlenecks will require richer multimodal benchmarks, standardized causal and structural evaluation, faithfulness-aware diagnostics for foundation models, and adaptive, weakly supervised learning, so that reasoning systems can be deployed transparently and reliably in open-world applications.

% \section*{Acknowledgements}

% \section*{Author information}
% \subsection{Authors and Affiliations}
% Ayushman Sarkar: Department of Computer Science and Engineering, Birbhum Institute
% of Engineering and Technology, Suri, 731101, West Bengal, India.

% \noindent
% Mohd Yamani Idna Idris: Faculty of Computer Science and Information Technology, Universiti Malaya, Kuala Lumpur, 50603, Malaysia.

% \noindent
% Zhenyu Yu: Faculty of Computer Science and Information Technology, Universiti
% Malaya, Kuala Lumpur, 50603, Malaysia.

% \subsection{Corresponding author}
% Correspondence to Zhenyu Yu (yuzhenyuyxl@foxmail.com).

\section*{Contributions}
\textbf{Ayushman Sarkar:} Investigation, Methodology, Writing (original draft), Writing (review and editing).

\textbf{Zhenyu Yu:} Conceptualization, Writing (review and editing).

\textbf{Mohd Yamani Idna Idris:} Writing (review and editing).

\section*{Ethics declarations}
The authors declare no competing interests.

\section*{Declaration of AI-Assisted Technologies}
During manuscript preparation, the authors used AI-based assistants to support language refinement and phrasing. All outputs were critically reviewed and edited by the authors, who take full responsibility for the final content.

% \appendix
% \section{My Appendix}
% Appendix sections are coded under \verb+\appendix+.

% \verb+\printcredits+ command is used after appendix sections to list 
% author credit taxonomy contribution roles tagged using \verb+\credit+ 
% in frontmatter.


\bibliographystyle{cas-model2-names}

% \bibliography{cas-refs}
\bibliography{main}

\end{document}